\documentclass[10pt,twocolumn,letterpaper]{article}

\PassOptionsToPackage{table}{xcolor}

\usepackage[final]{cvpr}      

\usepackage[most]{tcolorbox}   
\usepackage{multirow}          
\usepackage{pifont}            
\usepackage{comment}           

\definecolor{cvprblue}{rgb}{0.21,0.49,0.74}
\usepackage[pagebackref,breaklinks,colorlinks,allcolors=cvprblue]{hyperref}

\definecolor{bestgreen}{RGB}{217,242,222}
\definecolor{secondgreen}{RGB}{226,239,218}
\definecolor{pendingred}{RGB}{220,20,60}
\definecolor{lightgray}{RGB}{226,226,224}
\definecolor{lightviolet}{RGB}{237,231,246}
\definecolor{bestoverallviolet}{RGB}{230,220,255}

\newcommand{\best}[1]{\boldmath\textbf{#1}}
\newcommand{\second}[1]{\underline{#1}}
\newcommand{\third}[1]{#1}

\newcommand{\bestoverall}[1]{\cellcolor{lightgray}#1}

\newcommand{\cmark}{\ding{51}}
\newcommand{\PAR}[1]{\vskip 0pt \noindent{\bf #1~}}

\title{From Frames to Temporal Graphs: In-Context Egocentric Action Recognition with Vision-Language Models}
\newcommand*{\ourmodel}{TAG\xspace}

\author{Dominguez-Dager, Bessie\\
University of Alicante\\
\and
Gomez-Donoso, Francisco \\
University of Alicante\\
\and
Miguel Cazorla \\
University of Alicante\\
\and
Pollefeys, Marc \\
ETH Zurich, Microsoft\\
\and
Barath, Daniel  \\
ETH Zurich\\
\and
Bauer, Zuria  \\
ETH Zurich\\
}

\begin{document}
\twocolumn[{%
  \renewcommand\twocolumn[1][]{##1}
  \maketitle
\begin{center}
    \includegraphics[width=0.85\linewidth]{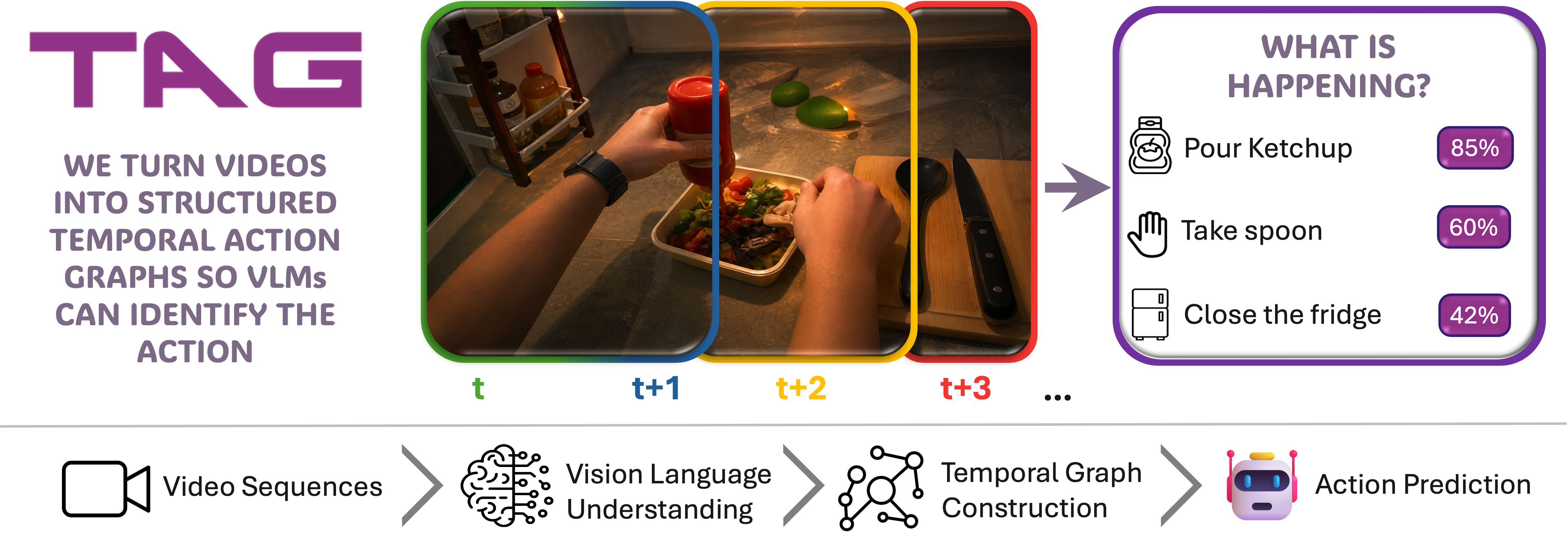}
    \captionof{figure}{The proposed \textbf{TAG} converts egocentric video into structured Temporal Action Graphs for action recognition. Video frames are processed through a VLM-based perception pipeline that extracts interaction descriptions and formalizes them into temporally ordered graph triplets encoding hand-object relations. The serialized graph is then provided to the VLM for closed-set action prediction, optionally augmented with labeled graph examples for in-context learning.}
    \label{fig:teaser}
    \vspace{3mm}
\end{center}

}]

\begin{abstract}
    Action reasoning in egocentric video requires capturing fine-grained transitions of hand-object interactions, a task where general-purpose Vision-Language Models (VLMs) often struggle when operating directly on raw pixels. We propose to decouple visual perception from symbolic reasoning by converting videos into Temporal Action Graphs. In a multi-stage prompting pipeline, we first generate dense natural language narratives over short temporal windows as a semantic bottleneck, then formalize them into structured, open-vocabulary graph representations. On the EGTEA and Epic-Kitchens-100 datasets, the symbolic representation unlocks efficient in-context learning: few-shot graph demonstrations yield substantial accuracy gains over zero-shot frame-based inference. Even in the zero-shot setting, graph-based reasoning remains competitive with pixel-based inference despite potential pretraining contamination favoring the latter. Across 11 open-weight VLMs from 6 model families ranging from 2B to 235B parameters, our findings indicate that current VLMs are more effective as symbolic reasoners than as direct visual observers. By projecting video into the language domain, we provide a scalable, fine-tuning-free alternative to end-to-end approaches that better leverages these models' latent reasoning strengths.
    The code will be made public.
\end{abstract}
\section{Introduction}
\label{sec:intro}

Understanding human actions in egocentric video is a foundational capability for applications ranging from robotic manipulation and assistive technologies to augmented reality. Unlike third-person video, egocentric footage is dominated by fine-grained hand-object interactions such as grasping, pouring, cutting, and placing. For these, the decisive visual evidence lies not in any single frame but in the transition between consecutive hand-object configurations.~\cite{egtea2018, ek100, grauman2022ego4d}. Recognizing such actions requires reasoning about temporal changes: what was being held, what is now being released, and what the hands are reaching for next.

Structured scene representations have proven effective for encoding relational information in both static and dynamic settings. Scene graphs, which model objects as nodes and their relationships as edges, have improved performance in visual question answering~\cite{johnson2015imagegraphs, xu2017scenegraph}, image generation~\cite{johnson2018image_gen_sg} and robotic planning~\cite{gu2024conceptgraphs, sayplan2023}. More recently, open-vocabulary 3D scene graphs have enabled language-grounded perception for navigation and manipulation~\cite{fun3Dsg, werby2024hovsg, koch2024open3dsg}, and dynamic scene graphs have extended these representations to capture temporal evolution~\cite{rosinol2020dsg, ji2020action_genome}. However, these graph-based methods have primarily targeted spatial understanding, object layouts and pairwise relations within a scene, rather than the temporal evolution of interaction states that characterizes egocentric actions.

Vision-Language Models (VLMs) have emerged as powerful general-purpose reasoning systems, capable of processing interleaved image and text inputs~\cite{alayrac2022flamingo, li2023blip, liu2023llava}. However, when applied to video-based action recognition, most VLMs face fundamental limitations: their training is overwhelmingly grounded in static image-text pairs, with limited or no explicit video supervision~\cite{liu2024tempcompass, mangalam2023egoschema}. When presented with a sequence of frames at inference time, these models must heuristically aggregate independent scene interpretations rather than reason over learned notions of motion or causality. The result is that fine-grained temporal dynamics, precisely those that distinguish taking from touching, or placing from releasing, are often lost in the dense, high-dimensional visual token stream.

We propose to decouple visual perception from symbolic reasoning. Rather than asking a VLM to perceive and reason over raw pixels, we use the model's visual capabilities to extract local interaction descriptions, which are then formalized into structured Temporal Action Graphs (\ourmodel). Each graph encodes the temporal evolution of hand-object interactions as a sequence of attributed triplets (source, relation, object), serializable as plain text. The VLM then reasons over this symbolic representation, operating in its native text domain, to predict the action label. This design exploits a key asymmetry: while VLMs often struggle as direct visual observers of dynamic scenes, they are highly effective symbolic reasoners when relational structure is stated directly in the graph~\cite{fatemi2024talk, wang2023nlgraph, gupta2023visual}. Since the serialized graph is plain text, it can itself serve as an in-context demonstration.

The main contributions can be summarized as follows:

\begin{itemize}
    \item We introduce \textbf{\ourmodel}, a framework that converts egocentric video into structured, open-vocabulary interaction graphs via multi-stage prompting, decoupling visual perception from symbolic reasoning without any training or fine-tuning.
    \item We show that reasoning over structured graphs outperforms direct frame-based inference in mean-class accuracy for all eleven evaluated VLMs, spanning 2B to 235B parameters, once graph demonstrations are provided in context, even under conditions where pretraining data contamination should favor pixel-based memorization.
    \item We demonstrate that graph serialization unlocks token-efficient in-context learning (ICL): prepending two labeled graph exemplars per class raises the average EGTEA MCA from $30.35\%$ (frames only) to $37.25\%$.
    \item We provide an extensive zero-shot and few-shot VLM evaluation for egocentric action recognition, covering eleven models on two major benchmarks (EGTEA and Epic-Kitchens-100), with systematic ablations over graph construction components, in-context example count, and input frame budget.
\end{itemize}




\section{Related work}
\label{sec:sota}

\textbf{Action recognition.}
EGTEA~\cite{egtea2018} and Epic-Kitchens-100~\cite{ek100} are the canonical fine-grained egocentric classification benchmarks, while Ego4D~\cite{grauman2022ego4d} dominates large-scale egocentric pretraining. Prior work on these benchmarks typically relies on egocentric-specialized pretraining: EgoVLP~\cite{lin2022egocentric}, LaViLa~\cite{zhao2023learning}, and EgoVideo~\cite{pei2025egovideo} fine-tune multimodal models on Ego4D narrations at substantial cost. The strongest zero-shot reference, GPT4Ego~\cite{dai2025gpt4ego}, composes off-the-shelf foundation tools (SAM~\cite{kirillov2023sam}, ChatGPT) in a multi-stage design but still freezes an egocentric-pretrained backbone (e.g., LaViLa) and selects classes by cosine similarity rather than generation. LLaVAction~\cite{qi2026llavaction} instead instruction-tunes a general-purpose VLM on the Epic-Kitchens-100 training split. We are not aware of published work that evaluates training-free, general-purpose, generative VLMs with in-context demonstrations on these benchmarks. Both datasets are public and likely partially represented in contemporary VLM training mixtures, a caveat that arguably favors pixel-based inference~\cite{dong-etal-2024-contamination}.

\PAR{Vision-language models.}
We evaluate eleven instruction-tuned VLMs from six open-weight families, spanning dense and mixture-of-experts (MoE) architectures from 2B to 235B parameters~\cite{li2025vlmsurvey}. Recent studies document severe limitations of VLMs on video understanding, especially in the egocentric setting~\cite{mangalam2023egoschema,plizzari2025egotempo,liu2024tempcompass,rodin2025easg}; we address these limitations through multi-stage prompting that surfaces temporal and relational structure to the VLM directly.

\PAR{Structured and language-based intermediate representations.}
Decoupling perception from reasoning has roots in neuro-symbolic AI~\cite{yi2018neural} and modern systems realize this through language as a zero-shot composition interface, e.g., Socratic Models for video QA~\cite{zeng2023socratic} or captions as inputs to frozen LLMs~\cite{yang2022empirical}. Closest in spirit, LLoVi~\cite{zhang2024llovi} captions short clips with a VLM and aggregates them with an LLM for long-form QA, mirroring our narrative stage but without graph formalization; pure language leaves compositional structure implicit. Graphs serve as compositional intermediates for visual reasoning, from static scene graphs~\cite{johnson2015imagegraphs} to spatio-temporal scene graphs~\cite{ji2020action_genome}. Recent work uses them as intermediate representations: EASG~\cite{rodin2024easg} annotates closed-vocabulary action scene graphs over Ego4D as a supervised target; SceneNet~\cite{taluzzi2025scenenet} prompts a proprietary VLM for scene graphs on HD-EPIC~\cite{hd-epic}; VideoMindPalace~\cite{huang2025building} serializes layered topological graphs; and SG-VLM~\cite{ma2025bridging} grounds reasoning with scene graphs via GroundingDINO~\cite{liu2024groundingdino} and SAM. All four extract graphs from the query video in a single grounding pass. Our two-stage narrative-then-graph pipeline instead deploys graphs as labeled few-shot demonstrations for fine-grained verb-noun classification, training-free.

\PAR{In-context learning with VLMs.}
ICL was popularized for LLMs by GPT-3~\cite{brown2020gpt3} and extended to multimodal inputs by Flamingo~\cite{alayrac2022flamingo}. Visual ICL adds a distinct bottleneck: each demonstration occupies hundreds to thousands of tokens, saturating context after a handful of examples~\cite{chen2023manipulating, jiang2024manyshot}. Serializing visual content as text compresses the demonstration budget and bypasses cross-modal interaction issues~\cite{zhou2024visualICL, yang2022empirical}; Zong et al.~\cite{zong2025vlicl} report that substituting text for images yields markedly steeper accuracy gains with the number of shots. We apply this strategy at the level of structured graphs rather than free-form captions. Closest to our setting, EILeV~\cite{yu2024eilev} elicits few-shot ICL on egocentric video by training BLIP-2 variants on Ego4D with carefully curated data; it documents that off-the-shelf VLMs fail to benefit from in-context examples on rare and out-of-distribution egocentric actions, the motivation we share. Our pipeline differs in three respects: demonstrations are text-serialized Temporal Action Graphs rather than raw video; the target task is fine-grained verb-noun classification rather than narration generation; and the pipeline is fully training-free.

\begin{figure*}[t]
    \centering
    \includegraphics[width=\linewidth]{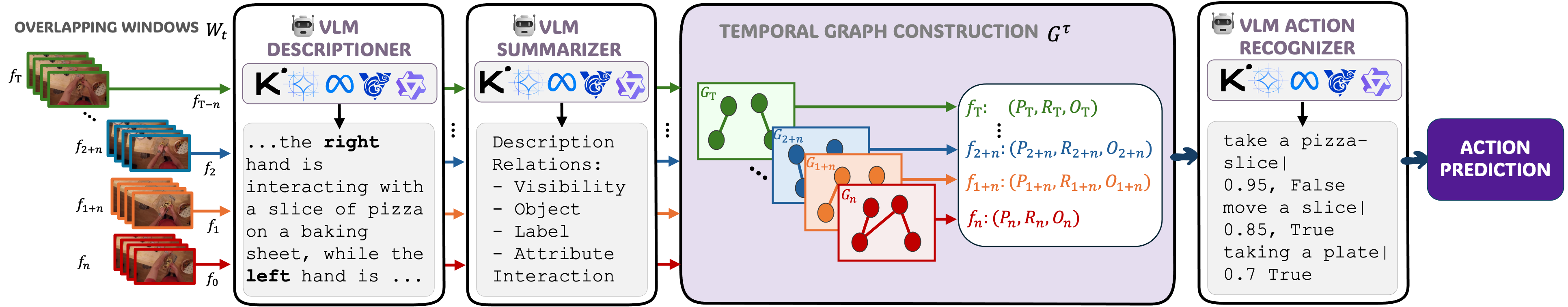}
    \caption{\textbf{Overview of the \ourmodel framework.} An egocentric video is processed through overlapping temporal windows $W_t$. For each window, a VLM generates a natural-language description of the dominant interaction, which is converted into a structured JSON graph. Window-level graphs are concatenated into a single Temporal Action Graph $\mathcal{G}^\tau$, serialized as temporally ordered triplets, and provided to the VLM for closed-set action prediction. The rightmost column illustrates the final prediction step with candidate actions and confidence scores. }
    \label{fig:pipeline}
\end{figure*}

\section{Method}
\label{sec:method}

We propose \ourmodel, a training-free framework that decouples visual perception from symbolic reasoning for egocentric action recognition. Given an egocentric video, \ourmodel distils pixel-level dynamics into a compact, text-serializable graph that a VLM can reason over in its native language domain. The framework has three stages (\cref{fig:pipeline}): temporal decomposition of the sampled frames into overlapping windows (\cref{sec:sliding_window}), neuro-symbolic mapping of each window into a local interaction graph via two-stage prompting (\cref{sec:neurosimbolic}), and temporal graph construction that assembles local graphs into a single Temporal Action Graph (\cref{sec:temporal_graph_construction}). The resulting graph is the input for closed-set action prediction, optionally augmented with labeled graph examples for ICL (\cref{sec:contextlearning}).

\PAR{Problem formulation.}
Let $V = \{f_t\}_{t=1}^T$ be the frames of an egocentric video. We uniformly sample $M$ frames to obtain a subset
\begin{equation}
    \mathbf{X} = (f_{\tau_1},\,f_{\tau_2},\ldots,\,f_{\tau_M}),
    \label{eq:sampling}
\end{equation}
where $\tau_i$ is the original temporal index of the $i$-th sampled frame. Given a closed action vocabulary $\mathcal{A} = \{a_1, \ldots, a_C\}$, the objective is to predict the ground-truth label $a^* \in \mathcal{A}$. A conventional VLM consumes the raw frames directly,
\begin{equation}
\label{eq:baseline}
\hat{a} = \Phi_{\mathrm{VLM}}(\mathbf{X},\, \mathcal{A}),
\end{equation}
whereas \ourmodel converts $\mathbf{X}$ into a Temporal Action Graph $\mathcal{G}^{\tau}$ (see \cref{sec:temporal_graph_construction}) and reasons over its text serialization as:
\begin{equation}
\label{eq:graphs-only}
\hat{a} = \Phi_{\mathrm{VLM}}(\mathrm{serialize}(\mathcal{G}^\tau),\, \mathcal{A}).
\end{equation}
Comparing \cref{eq:baseline,eq:graphs-only} under the same VLM and  sampling protocol isolates the effect of the intermediate representation from the capacity of the perception module.

\vspace{-1mm}
\subsection{Temporal decomposition via overlapping sliding windows}
\label{sec:sliding_window}
\vspace{-1mm}

Egocentric actions are defined by transitions between hand-object configurations: an action such as ``take'' is identified by the shift from reaching to grasping rather than by any single static frame. To expose such local dynamics while keeping each extraction step tractable, we partition $\mathbf{X}$ into overlapping temporal windows.

\PAR{Window definition.}
Graph construction uses the $M{-}2$ interior frames: we drop the first and last sampled frame, which are least likely to carry the interaction in trimmed segments, which also saves two windows of extraction. Given a maximum window size $n$, for each anchor $t \in \{3, \dots, M{-}1\}$ we define
\begin{equation}
    W_t = (f_{\tau_{\max(2,\,t-n+1)}}, \ldots, f_{\tau_t}),
\end{equation}
yielding $M{-}3$ windows ($13$ for $M{=}16$). The first $n{-}2$ grow from two frames; all subsequent ones have exactly $n$ frames, with stride one (overlap $n{-}1$). We fix $n = 4$, so each window spans two to four consecutive sampled frames.

\PAR{Anchor principle.}
Each window $W_t$ is anchored to its final frame $f_{\tau_t}$: although the VLM sees the full window, symbolic extraction (\cref{sec:neurosimbolic}) is constrained to describe the dominant interaction visible at $f_{\tau_t}$. The preceding $n{-}1$ frames serve as local context to disambiguate the ongoing interaction (e.g., releasing vs.\ touching). This guarantees exactly one interaction snapshot per interior frame (except $f_{\tau_2}$, used only as context for the first window), avoiding redundancy while preserving temporal coverage.

\vspace{-1mm}
\subsection{Neuro-symbolic mapping: from pixels to local graphs}
\label{sec:neurosimbolic}
\vspace{-1mm}

Each window $W_t$ is converted into a structured local graph $G_t$ through a two-stage prompting strategy. The key insight is that decomposing the visual-to-symbolic translation into an intermediate natural-language description and a subsequent formalization step reduces symbolic hallucinations and improves grounding, consistent with findings that expanded intermediate-reasoning budgets benefit complex extraction tasks~\cite{wei2022chain, yao2023tree}.

\PAR{Stage I: semantic description.}
Given a temporal window $W_t$, the VLM generates a free-form natural-language narrative
\begin{equation}
\mathcal{N}_t = \Phi_{\mathrm{VLM}}(W_t,\, p_{\mathrm{desc}}),
\end{equation}
where the description prompt $p_{\mathrm{desc}}$ instructs the model to (i)~identify the action performed by the hands or, if hands are not visible or not informative, by the camera wearer; (ii)~specify the roles of the left/right hands when visible; and (iii)~disambiguate visually similar objects by grounding them in the visual evidence. The narrative $\mathcal{N}_t$ acts as a semantic bottleneck that forces the VLM to consolidate visual evidence into a coherent linguistic summary before any symbolic structure is imposed.

\PAR{Stage II: symbolic extraction.}
The narrative is then parsed into a structured graph. Crucially, this stage operates on $\mathcal{N}_t$ alone, without re-processing the original frames, ensuring that all visual grounding is mediated through the bottleneck:
$G_t = \Phi_{\mathrm{VLM}}(\mathcal{N}_t,\, p_{\mathrm{struct}})$,
where $p_{\mathrm{struct}}$ is the structuring prompt.

\PAR{Local graph schema.}
The output $G_t$ is a set of up to $K_t$ interaction triplets:
$G_t = \{g_{t,k}\}_{k=1}^{K_t}$ with $g_{t,k} = (s_{t,k},\, r_{t,k},\, o_{t,k})$,
where the three components are defined as follows.
\begin{itemize}[nosep]
    \item \textbf{Source:} $s_{t,k} \in \mathcal{S}$, the interaction agent, is drawn from a fixed vocabulary
        \[
        \begin{aligned}
        \mathcal{S} = \{\,&\texttt{hand\_left},\;\texttt{hand\_right},\\
        &\texttt{hand\_both},\;\texttt{camera\_wearer}\,\}.
        \end{aligned}
        \]
    \item \textbf{Relation:} $r_{t,k} \in \mathcal{R}$, an open-vocabulary verb in continuous form (e.g., holding, cutting, reaching), describing the interaction type.
    \item \textbf{Object:} $o_{t,k} = (\alpha_{t,k},\,\ell_{t,k})$ where $\ell_{t,k}$ is a one-word object label and $\alpha_{t,k}$ is a one-word attribute a salient property (color, state, material, or function). The attribute-label pair forms an open-vocabulary attributed object.
\end{itemize}

\PAR{Schema constraints.}
Four constraints keep $G_t$ compact and temporally localized: (i)~\texttt{hand\_both} is used only when the left and right hands cooperate on the same object with the same interaction; (ii)~when hand interactions are present, camera-wearer relations are excluded, prioritizing active manipulation over generic ego-motion; (iii)~only the interaction state at the terminal frame is included, even if the narrative references earlier frames; (iv)~at most one relation per interaction source, preventing duplicates for the same agent.


\vspace{-1mm}
\subsection{Temporal Action Graph construction}
\label{sec:temporal_graph_construction}
\vspace{-1mm}

The window-level graphs are assembled into a directed attributed graph $\mathcal{G}^{\tau} = (\mathcal{V}^{\tau},\, \mathcal{E}^{\tau})$ that captures both the semantic content and the temporal evolution of the video. The node set aggregates all sources and attributed objects observed across the video,
\begin{equation}
\mathcal{V}^\tau = \bigcup_{t=3}^{M-1}\bigcup_{k=1}^{K_t} \{s_{t,k},\, o_{t,k}\},
\end{equation}
and each edge extends a local triplet with a temporal index:
\begin{equation}
\mathcal{E}^\tau = \bigcup_{t=3}^{M-1}\bigcup_{k=1}^{K_t} \bigl\{(s_{t,k},\, r_{t,k},\, o_{t,k},\, t)\bigr\}.
\end{equation}
The temporal index $t$ preserves ordering, allowing the graph to represent both persistence (repeated relations) and progression (changes in relation or object).

\PAR{Serialization.}
For inference, $\mathcal{G}^\tau$ is serialized into a temporally ordered sequence of triplets,
\begin{equation}
    \textsc{serialize}(\mathcal{G}^{\tau}) = \bigl[\,(s_{t,k},\; r_{t,k},\; o_{t,k})\;\big|_{t=3}^{M-1}\,\bigr],
    \label{eq:serialize}
\end{equation}
where triplets within a timestamp are ordered by source priority (left hand, right hand, both hands, camera wearer) and timestamps follow chronological order. This deterministic ordering yields a unique, reproducible string that the VLM processes directly, without a dedicated graph encoder.

\vspace{-1mm}
\subsection{Action prediction and in-context learning}
\label{sec:contextlearning}
\vspace{-1mm}

\Cref{eq:baseline,eq:graphs-only} define the \emph{frames-only} and \emph{graphs-only} settings tested in our experiments. The third setting, \emph{Graphs + ICL}, prepends $N\!=\!k|\mathcal{A}|$ labeled graph exemplars, $k$ per candidate class, to the query graph:
\begin{equation}
\small 
\hat{a} = \Phi_{\mathrm{VLM}}\!\left(
  \bigl(\mathrm{serialize}(\mathcal{G}_n^\tau),\, a_n\bigr)_{n=1}^{N},\,
  \mathrm{serialize}(\mathcal{G}^\tau),\,
  \mathcal{A}
\right).
\end{equation}
Graph serialization is what makes ICL practical for video: a class-balanced pool over EGTEA's $106$ classes is $212$ short text strings at $k\!=\!2$, whereas the visual equivalent is $3{,}392$ frames, which at a few hundred image tokens each would saturate the context window~\cite{jiang2024manyshot}. Projecting video into a symbolic format also aligns the task with the VLM's pre-trained pattern matching over text sequences, bypassing the lack of explicit cross-image relational reasoning in standard VLM training~\cite{zhou2024visualICL}.

\section{Experiments}
\label{sec:exp}

\PAR{Datasets and evaluation protocols.}
On \textit{EGTEA}~\cite{egtea2018} (106 action classes), we follow the official protocol on split~1 (8{,}299/2{,}022 train/test) and report  mean-class-accuracy (MCA), Top-1, Top-3, and Top-5. On \textit{Epic-Kitchens-100}~\cite{ek100} (97 verbs, 300 nouns, 3{,}806 composed action classes; 67{,}217/9{,}668 train/val), we evaluate on val and report Top-1/Top-5 for verb, noun, and action, with an action counted correct only when both verb and noun match. The training split is used solely as the ICL retrieval pool.

\PAR{Evaluated VLMs.}
We evaluate eleven open-weight VLMs from six families, spanning 2B--235B parameters, across dense and MoE architectures, in three regimes by total count: \emph{small} ($<\!20$B), \emph{medium} (20--100B), and \emph{large} ($>\!100$B). Checkpoints that do not fit in GPU memory at full precision are quantized and marked $\ast$.
Detailed descriptions can be found in the Supplementary, \cref{app:models}.

\PAR{Inference setup.}
All experiments use vLLM~\cite{vllm} on up to four NVIDIA L40S GPUs with greedy decoding (temperature $0$) and reasoning modes disabled. Every modality starts from the same 16 uniformly sampled frames per clip and the same closed candidate set of action labels: \emph{(i) Frames only:} the 16 frames and candidate set are passed to the VLM. \emph{(ii) Graphs only:} the frames feed the TAG construction of \cref{sec:method}, which uses the 14 interior frames, and the serialized graph $\mathcal{G}^{\tau}$ replaces them as input. \emph{(iii) Graphs + ICL:} the graphs-only prompt is prefixed with labeled graph examples from the training split. The candidate set, response schema, and ranking semantics are identical across all three. The graph modalities therefore see two frames fewer than the baseline; since frames-only accuracy grows with the frame budget (\cref{tab:ablation_frames_4_vs_16}), this asymmetry favors the baseline.

\PAR{Prompting.}
A fixed template across modalities and models contains (i)~a task instruction, (ii)~the input, (iii)~the closed candidate set, and (iv)~a constrained JSON response schema. For EGTEA, candidates are \texttt{action\_id: action\_label} pairs and the model returns the five most plausible actions. For Epic-Kitchens-100, verb and noun vocabularies are listed separately and the model returns five \texttt{verb:noun} predictions; pairs with valid verb and noun but no annotated action class are counted for verb and noun Top-1/Top-5 only. Responses are accepted only if they contain exactly five distinct predictions from the candidate set with confidences in $[0,1]$ (used only for ranking). Full templates are in the Supplementary, \cref{app:action_prompt}.

\PAR{ICL pool.}
The pool is built once per seed from the training split and held fixed across models and queries, so differences are attributable to the VLM and the input representation. For EGTEA we use a class-balanced pool with $k\!\in\!\{1,2\}$ examples per class. For Epic-Kitchens-100, demonstrations for all $3{,}806$ classes do not fit in context, so the pool covers only the $m\!\in\!\{50,100\}$ most frequent training classes. The candidate set stays the full vocabulary, so the model must still predict classes for which it has seen no demonstration.

\vspace{-1mm}
\subsection{Main results}
\vspace{-1mm}

In all tables we \colorbox{lightgray}{\makebox[1cm][c]{highlight}}, \best{bold}, and \underline{underline} the best overall, individual best, and second-best entries, respectively, and report ICL results as mean $\pm$ SEM (standard error of the mean) over 5 seeds.

\begin{table*}
  \caption{\textbf{Main results on EGTEA.} MCA and Top-1 on the split-1 test set ($106$ action classes) for all eleven backbones, ordered by total parameters. ICL entries use $k\!=\!2$ examples per class and are mean $\pm$ SEM over $5$ demonstration draws. The last block macro-averages each modality across all models.}
  \label{tab:mca_top1_all_models}
  \vspace{-3mm}
  \centering
  \small
  \setlength{\tabcolsep}{10pt}
  \begin{tabular}{lccccccc}
    \toprule
    & & \multicolumn{2}{c}{Frames only}
    & \multicolumn{2}{c}{Graphs only}
    & \multicolumn{2}{c}{Graphs + ICL} \\
    \cmidrule(lr){3-4}
    \cmidrule(lr){5-6}
    \cmidrule(lr){7-8}
    Model & Size
    & MCA & Top-1
    & MCA & Top-1
    & MCA & Top-1 \\
    \midrule
    Qwen3.5-2B & 2B
    & \second{25.7} & \third{32.8}
    & \third{20.6} & \second{35.0}
    & \bestoverall{\best{32.2 $\pm$ 0.2}} & \bestoverall{\best{38.2 $\pm$ 0.2}} \\
    Gemma-3-4B & 4B
    & \third{8.7} & \third{15.0}
    & \bestoverall{\best{14.2}} & \bestoverall{\best{32.3}}
    & \second{12.1 $\pm$ 0.2} & \second{23.8 $\pm$ 0.4} \\
    GLM-4.6V-Flash & 9B
    & \third{27.1} & \third{39.7}
    & \second{31.6} & \second{45.0}
    & \bestoverall{\best{36.6 $\pm$ 0.4}} & \bestoverall{\best{46.8 $\pm$ 0.3}} \\
    Kimi-VL-A3B & 16B
    & \third{20.0} & \third{34.0}
    & \second{21.0} & \second{37.9}
    & \bestoverall{\best{29.0 $\pm$ 0.5}} & \bestoverall{\best{41.4 $\pm$ 0.4}} \\
    \midrule
    Gemma-3-27B & 27B
    & \third{23.1} & \third{34.7}
    & \bestoverall{\best{26.8}} & \bestoverall{\best{40.6}}
    & \second{24.9 $\pm$ 0.4} & \second{38.4 $\pm$ 0.5} \\
    Qwen3-VL-32B & 32B
    & \third{34.6} & \third{47.6}
    & \second{37.2} & \second{50.2}
    & \bestoverall{\best{42.4 $\pm$ 0.4}} & \bestoverall{\best{53.4 $\pm$ 0.2}} \\
    Qwen3.5-35B-A3B & 35B
    & \bestoverall{\second{50.5}} & \bestoverall{\best{57.3}}
    & \third{45.3} & \third{52.9}
    & \best{53.1 $\pm$ 0.5} & \second{54.0 $\pm$ 0.4} \\
    \midrule
    GLM-4.6V* & 106B
    & \third{35.8} & \third{48.5}
    & \second{42.4} & \second{52.0}
    & \bestoverall{\best{47.7 $\pm$ 0.3}} & \bestoverall{\best{54.5 $\pm$ 0.2}} \\
    Llama-4-Scout* & 109B
    & \third{17.4} & \third{31.0}
    & \second{25.0} & \second{40.2}
    & \bestoverall{\best{29.7 $\pm$ 0.3}} & \bestoverall{\best{44.4 $\pm$ 0.1}} \\
    Qwen3.5-122B* & 122B
    & \second{51.7} & \best{58.1}
    & \third{47.2} & \third{53.6}
    & \bestoverall{\best{56.1 $\pm$ 0.5}} & \bestoverall{\second{57.6 $\pm$ 0.4}} \\
    Qwen3-VL-235B* & 235B
    & \third{39.2} & \third{49.7}
    & \second{40.6} & \second{52.3}
    & \bestoverall{\best{46.1 $\pm$ 0.5}} & \bestoverall{\best{56.2 $\pm$ 0.3}} \\
    \midrule
    Overall & --
    & \third{30.4} & \third{40.8}
    & \second{32.0} & \second{44.7}
    & \bestoverall{\best{37.3}} & \bestoverall{\best{46.2}} \\
    \bottomrule
  \end{tabular}%
  \vspace{-1mm}
\end{table*}

\textbf{EGTEA.}
\emph{Graphs + ICL} improves MCA over frames-only on all eleven backbones (\cref{tab:mca_top1_all_models}) and Top-1 on $9$ of $11$; on GLM-4.6V-Flash it beats frames by $9.5$ MCA, supporting the view that the symbolic graph format is a memory-efficient interface for in-context conditioning, where raw visual tokens would saturate the context window. Zero-shot graphs are less uniform: they improve Top-1 on $9$ of $11$ backbones and MCA on $8$, but lose to frames on both metrics on Qwen3.5-35B and Qwen3.5-122B$^{*}$, the two strongest frame baselines, so on those models the gain comes from the demonstrations rather than the graph alone. Llama-4-Scout$^{*}$ in particular has an unusually weak frame baseline ($17.4$ MCA) yet reaches $29.7$ with graphs + ICL, suggesting graphs partially compensate for limited temporal alignment in pretraining.

\textbf{Epic-Kitchens-100.}
On Epic-Kitchens-100 (\cref{tab:ek100_main_results}), \emph{Graphs + ICL is strongest overall}: it beats frames-only on every macro-average metric, with the largest gain on the strongest backbone (GLM-4.6V$^{*}$: Action Top-1 $10.60\!\to\!14.02$, Top-5 $17.68\!\to\!26.69$). SEMs are small relative to the gaps ($\pm 0.15$ on Action Top-1 for GLM-4.6V$^{*}$), so the improvement is stable across examples rather than an artifact of a particular shot. Notably, the pool demonstrates only $100$ of the $3{,}806$ action classes, while predictions range over the full vocabulary.

\begin{table*}
  \caption{\textbf{Main results on Epic-Kitchens-100.} Top-1 and Top-5 accuracy for verb, noun, and composed action across three VLMs. Graphs + ICL uses one in-context example for each of the $100$ most frequent training classes, while all modalities predict over the full label vocabulary; \textit{Overall} reports the macro-average over the three models per modality. ICL entries are mean $\pm$ SEM over $5$ seeds.}
  \label{tab:ek100_main_results}
  \vspace{-3mm}
  \centering
  \small
  \setlength{\tabcolsep}{5pt}
  \resizebox{\textwidth}{!}{
  \begin{tabular}{llcccccc}
    \toprule
    \multirow{2}{*}{Model} & \multirow{2}{*}{Input Modality}
    & \multicolumn{3}{c}{Top-1 Acc.}
    & \multicolumn{3}{c}{Top-5 Acc.} \\
    \cmidrule(lr){3-5} \cmidrule(lr){6-8}
    & & Verb & Noun & Action & Verb & Noun & Action \\
    \midrule

    \multirow{3}{*}{GLM-4.6V-Flash}
      & Frames only   & 24.54 & 23.71 & 9.91 & 26.53 & 35.44 & 13.83 \\
      & Graphs only   & 18.76 & 24.09 & 6.68 & 26.64 & 38.49 & 12.86 \\
      & \bestoverall{Graphs + ICL}  
      & \bestoverall{\best{28.59 $\pm$ 0.40}} 
      & \bestoverall{\best{27.22 $\pm$ 0.12}} 
      & \bestoverall{\best{10.65 $\pm$ 0.10}} 
      & \bestoverall{\best{40.42 $\pm$ 0.65}} 
      & \bestoverall{\best{40.95 $\pm$ 0.28}} 
      & \bestoverall{\best{20.10 $\pm$ 0.22}} \\
    \midrule

    \multirow{3}{*}{Kimi-VL-A3B}
      & Frames only   & \best{32.09} & 17.67 & 6.33 & \best{38.29} & 28.40 & 12.48 \\
      & Graphs only   & 25.77 & 22.04 & 7.26 & 29.34 & \best{36.05} & 12.88 \\
      & \bestoverall{Graphs + ICL}  
      & \bestoverall{27.59 $\pm$ 0.33} 
      & \bestoverall{\best{24.00 $\pm$ 0.08}} 
      & \bestoverall{\best{9.09 $\pm$ 0.15}} 
      & \bestoverall{33.30 $\pm$ 0.14} 
      & \bestoverall{33.58 $\pm$ 0.17} 
      & \bestoverall{\best{13.93 $\pm$ 0.11}} \\
    \midrule

    \multirow{3}{*}{GLM-4.6V*}
      & Frames only   & 27.18 & 26.22 & 10.60 & 32.73 & 43.12 & 17.68 \\
      & Graphs only   & 24.88 & 25.81 & 9.01 & 34.43 & 41.93 & 16.32 \\
      & \bestoverall{Graphs + ICL}  
      & \bestoverall{\best{34.58 $\pm$ 0.21}} 
      & \bestoverall{\best{29.57 $\pm$ 0.13}} 
      & \bestoverall{\best{14.02 $\pm$ 0.15}} 
      & \bestoverall{\best{47.96 $\pm$ 0.29}} 
      & \bestoverall{\best{46.66 $\pm$ 0.05}} 
      & \bestoverall{\best{26.69 $\pm$ 0.15}} \\
    \midrule

    \multirow{3}{*}{Overall}
      & Frames only   & 27.94 & 22.53 & 8.95 & 32.52 & 35.65 & 14.66 \\
      & Graphs only   & 23.14 & 23.98 & 7.65 & 30.14 & 38.82 & 14.02 \\
      & \bestoverall{Graphs + ICL}  
      & \bestoverall{\best{30.25}} 
      & \bestoverall{\best{26.93}} 
      & \bestoverall{\best{11.25}} 
      & \bestoverall{\best{40.56}} 
      & \bestoverall{\best{40.40}} 
      & \bestoverall{\best{20.24}} \\
    
    \bottomrule
  \end{tabular}}
  \vspace{-4mm}
\end{table*}

\emph{ICL primarily recovers the verb gap.} The largest single jump over \emph{graphs only} comes in verb accuracy ($23.14\!\to\!30.25$ Top-1, macro-average). Since examples are themselves graphs, we read this not as additional visual evidence but as supplying a decoding pattern from serialized graph to closed-set label, which is the bottleneck when graphs are used alone. Kimi-VL-A3B is the exception: its frames-only verb accuracy ($32.09$) is the table's highest, yet ICL still improves action ($6.33\!\to\!9.09$) and noun ($17.67\!\to\!24.00$) accuracies.

\subsection{ICL scaling analysis}

\begin{figure*}[t]
  \centering
  \includegraphics[width=\linewidth]{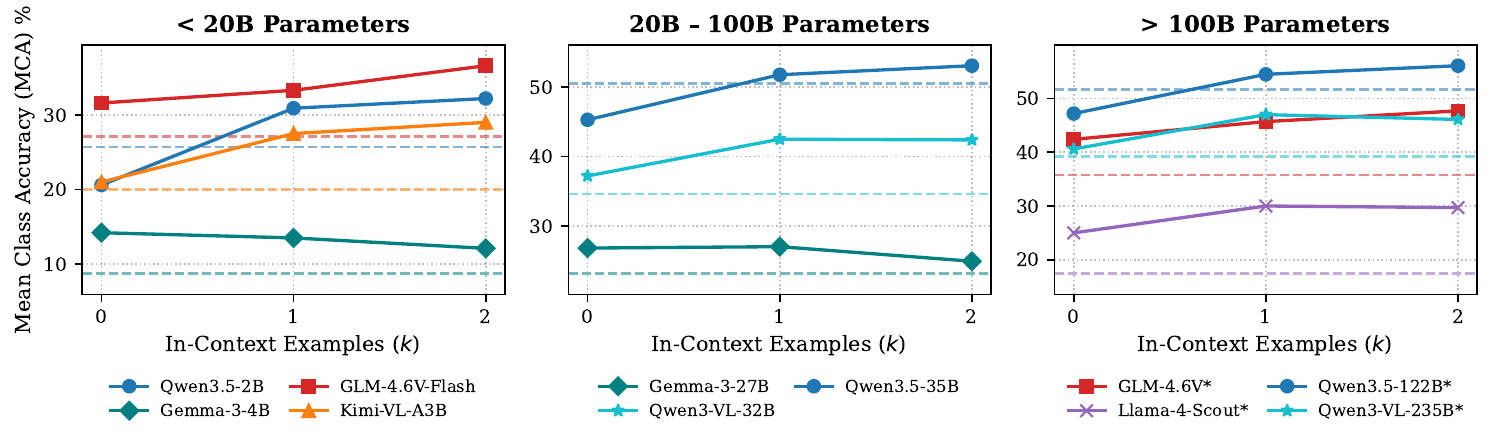}
  \vspace{-6mm}
    \caption{\textbf{ICL scaling on EGTEA} across different model parameter regimes (MCA). The horizontal dashed lines indicate the zero-shot raw frame baseline for each respective model. This visualizes the crossover point: while zero-shot graphs occasionally underperform dense frame inputs, scaling to $k \ge 1$ graph examples consistently eclipses the frames-only bounds across every tested architecture.}
  \label{fig:icl_scaling_egtea}
\end{figure*}

\Cref{fig:icl_scaling_egtea} sweeps the number of graph demonstrations $k\!\in\!\{0,1,2\}$  per class on EGTEA, grouped by parameter regime; we average MCA over $5$ independent demonstration draws.\footnote{Per-seed SEMs and Top-$\{1,3,5\}$ accuracies are reported in the {Supplementary \cref{app:supp_per_model}}.} For every family except Gemma-3, a single demonstration ($k\!=\!1$) already surpasses both graphs-only and frames-only; a second helps Qwen3.5, GLM-4.6V, and Kimi-VL but slightly hurts Qwen3-VL and Llama-4-Scout$^{*}$, so the optimal $k$ is model-dependent. Gemma-3 inverts this pattern. Although its graphs-only baseline exceeds frames-only, demonstrations do not improve on it; whether demonstrations help is known to vary sharply across backbones~\cite{zong2025vlicl}, and for this family the added prompt length appears to outweigh their instructional signal.

\begin{figure*}[t]
  \centering
  \includegraphics[width=\linewidth]{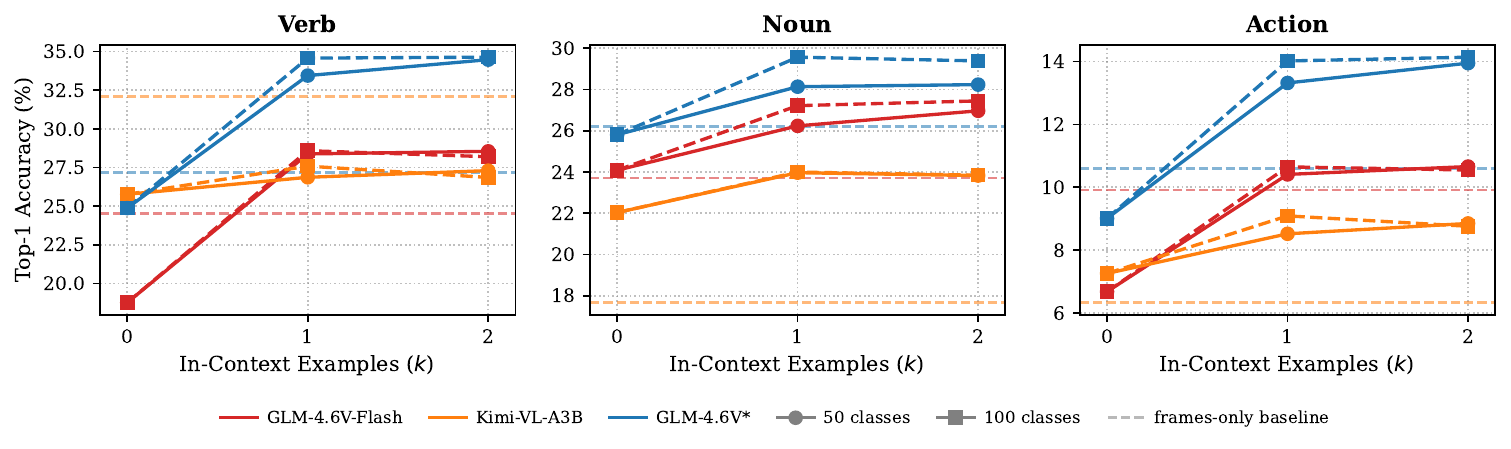}
  \vspace{-7mm}
    \caption{\textbf{ICL scaling on Epic-Kitchens-100} across Verb, Noun, and Action (Top-1). Each panel reports Top-1 accuracy versus the number of graph-based in-context examples per class (\textbf{0} corresponds to the graphs-only setting, i.e., no in-context demonstrations). Solid lines denote a pool covering the 50 most frequent classes; dashed lines, 100. Long-dashed horizontals mark the frames-only baseline.}
  \label{fig:icl_scaling_ek100}
  \vspace{-2mm}
\end{figure*}

\begin{table}[t]
\centering
\small
\caption{\textbf{TAG construction ablation on EGTEA.} Joint ablation over overlapping windows (\textit{Window}), multi-stage prompting (\textit{MS-Prompt.}), and node attributes (\textit{Attr.}) on two Qwen3.5 backbones. The last row of each block is the full configuration.}
\label{tab:ablation_attr_sliding_cot_joint}
\resizebox{\columnwidth}{!}{%
\begin{tabular}{lccc cccc}
\toprule
Model & Window & MS-Prompt. & Attr. & MCA & Top-1 & Top-3 & Top-5 \\
\midrule


\multirow{4}{*}{Qwen3.5-35B}
&   & \cmark & \cmark & 40.83 & 48.62 & 68.10 & 75.22 \\
& \cmark &   & \cmark & 41.71 & 50.00 & 70.33 & 77.89 \\
& \cmark & \cmark &   & 43.30 & 50.94 & 70.72 & 77.65 \\
& \cmark & \cmark & \cmark & \best{45.34} & \best{52.87} & \best{73.99} & \best{80.27} \\
\midrule

\multirow{4}{*}{Qwen3.5-122B*}
&   & \cmark & \cmark & 41.53 & 49.11 & 70.77 & 79.43 \\
& \cmark &   & \cmark & 42.79 & 50.40 & 72.40 & 80.81 \\
& \cmark & \cmark &   & 44.08 & 51.83 & 74.00 & 81.45 \\
& \cmark & \cmark & \cmark & \best{47.17} & \best{53.61} & \best{76.16} & \best{83.93} \\
\bottomrule
\end{tabular}%
}
\vspace{-4mm}
\end{table}

\begin{table}[t]
\centering
\small
\caption{\textbf{Frame-budget ablation on EGTEA.} \textit{Frames only} accuracy at $4$ vs.\ $16$ input frames; $4$ matches the per-window count used during graph construction.}
\label{tab:ablation_frames_4_vs_16}
  \vspace{-1.5mm}
\setlength{\tabcolsep}{5pt}
\begin{tabular}{llcccc}
\toprule
Model & Frames & MCA & Top-1 & Top-3 & Top-5 \\
\midrule
\multirow{2}{*}{Qwen3.5-2B} & 4  & 22.21 & 31.31 & 49.55 & 55.89 \\
 & 16 & \best{25.65} & \best{32.84} & \best{49.60} & \best{56.53} \\
\midrule
\multirow{2}{*}{Qwen3.5-35B} & 4 & 46.31	& 55.34	 & 70.28 & 76.66 \\
 & 16 & \best{50.53} & \best{57.27} & \best{74.73} & \best{80.02} \\
\midrule
\multirow{2}{*}{Qwen3.5-122B*} & 4  & 44.93 & 52.67 & 70.57 & 78.09 \\
 & 16 & \best{51.73} & \best{58.06} & \best{74.83} & \best{80.81} \\
\bottomrule
\end{tabular}
\vspace{-5.4mm}
\end{table}
\Cref{fig:icl_scaling_ek100} sweeps both the number of classes the demonstration pool covers ($50$ or $100$) and the number of graph examples per class on EK100. One example per class consistently improves over graphs-only and frames-only across verb, noun, and action; a second example helps at $50$ classes but not at $100$, where the longer prompt can slightly hurt the smaller models. The best trade-off is one example per class over the $100$ most frequent classes. Graph ICL helps, but does not scale monotonically with $k$.

\subsection{Ablations}

\PAR{TAG construction.}
We ablate three orthogonal design choices on EGTEA (\cref{tab:ablation_attr_sliding_cot_joint}, two Qwen3.5 backbones): (i)~overlapping temporal windows, (ii)~multi-stage prompting (MS-Prompt.), and (iii)~node attributes. Each row removes one factor from the full configuration in row~4. Sliding windows are the largest single contributor on both backbones ($+4.51$/$+5.64$ MCA on Qwen3.5-35B / Qwen3.5-122B$^{*}$); followed by MS-Prompt ($+3.63$/$+4.38$); attributes help every metric ($+2.04$/$+3.09$ MCA), with a larger gain on the larger model.

\begin{figure*}[t]
    \centering
    \includegraphics[width=\linewidth, height=10cm]{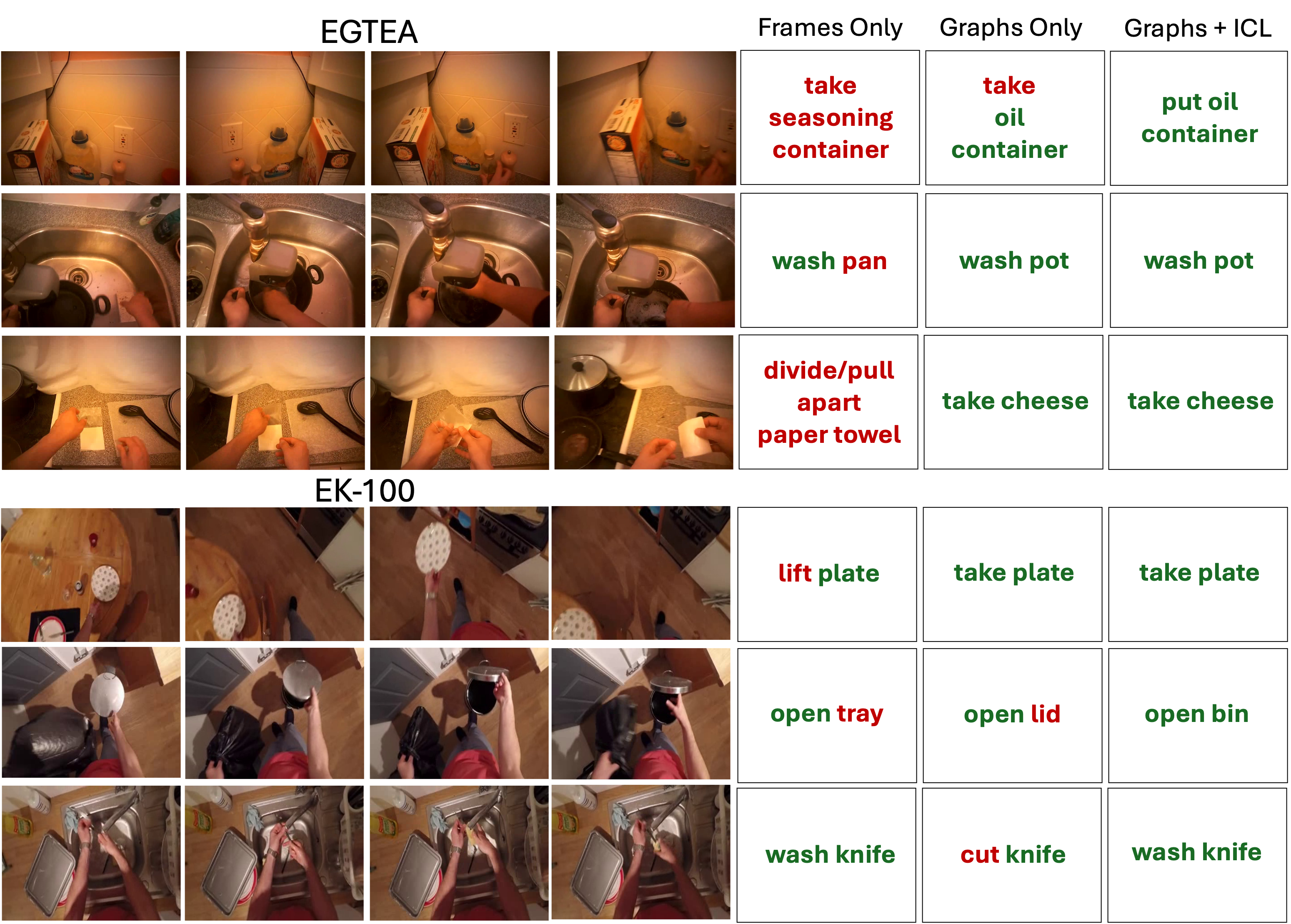}
    \caption{\textbf{Qualitative results of \ourmodel on EGTEA and Epic-Kitchens-100.}  For each clip (rows), we show four sampled frames and the predicted label under \emph{Frames only}, \emph{Graphs only}, and \emph{Graphs + ICL}, with correct predictions in \textbf{\textcolor{green!50!black}{green}} and errors in \textbf{\textcolor{red}{red}}. The temporal graph $\mathcal{G}^{\tau}$ already resolves several verb-level errors made under \emph{Frames only}, and adding in-context examples further disambiguates fine-grained verb and noun confusions that persist in the \emph{Graphs only} variant. Results obtained with Qwen3.5-122B* on EGTEA and GLM-4.6V* on Epic-Kitchens-100.}
    \label{fig:qualitative}
\end{figure*}

\PAR{Frame budget in \emph{Frames only}.}
To rule out a context-size explanation, we compare \emph{frames-only} at 4 vs.\ 16 input frames, with 4 matching the per-window count used during graph construction (\cref{tab:ablation_frames_4_vs_16}). More frames consistently help every metric and model, so the gains from graph-based prompting (despite its 4-frame windows), cannot be attributed to seeing fewer frames at once, but to the structured temporal abstraction itself. This is consistent with EgoVideo, which reports saturation only at 32 frames~\cite{pei2025egovideo}.

\Cref{fig:qualitative} illustrates two recurring failure modes for \emph{Frames only}: temporal ambiguity between near-symmetric verbs, e.g., \emph{take} and \emph{put}, and unstable noun predictions under partial object occlusion. Confusions are progressively resolved from \emph{Frames only} to \emph{Graphs only} to \emph{Graphs + ICL}, though \emph{Graphs only} also introduces an error when the underlying VLM misreads hand-object interactions within a short window (e.g.\ \emph{cut} vs.\ \emph{wash}), which ICL subsequently corrects. In the end, the graph representation is only as reliable as the per-window perception that feeds it.

\section{Conclusions}
\label{sec:conc}

We demonstrate that converting an egocentric video into Temporal Action Graphs shifts action recognition from the visual stream into the language domain. By decoupling perception from reasoning, this text-serializable bottleneck enables competitive zero-shot inference and unlocks few-shot in-context learning that remains impractical with raw video. While end-to-end compute scaling may eventually solve video understanding natively, structured language is a pragmatic bridge today: until video pretraining paradigms induce robust temporal reasoning, explicit symbolic representations let us recruit a VLM's latent linguistic strengths to compensate for its visual limitations.

\PAR{Limitations.}
Accuracy remains below specialized egocentric models fine-tuned on in-domain data, and is bounded by the quality of the Stage I narrative, since the symbolic bottleneck cannot recover what the perception step misses. The pipeline also incurs higher latency than single-pass models, requiring $\sim\!27$ forward passes per clip. A direct multiplier comparison is misleading: Stage I calls receive $4$ frames and Stage II calls text alone, against $16$ images in the frames-only call. Graph construction is furthermore a one-time cost per clip: the same graph serves every ICL configuration, so increasing $k$ adds text tokens but no VLM calls. We defer a detailed latency analysis to future work; implementation details are in the Supplementary Material.

{
    \small
    \bibliographystyle{ieeenat_fullname}
    \bibliography{main}

@String(CVPR= {IEEE Conf. Comput. Vis. Pattern Recog.})

@String(ECCV= {Eur. Conf. Comput. Vis.})

@String(NIPS= {Adv. Neural Inform. Process. Syst.})

@String(AAAI = {AAAI})

@String(CVPRW= {IEEE Conf. Comput. Vis. Pattern Recog. Worksh.})

@String(CVPR  = {CVPR})

@String(ECCV  = {ECCV})

@String(NIPS  = {NeurIPS})

@String(CVPRW= {CVPRW})

@inproceedings{gu2024conceptgraphs,
  title={Conceptgraphs: Open-vocabulary 3d scene graphs for perception and planning},
  author={Gu, Qiao and Kuwajerwala, Ali and Morin, Sacha and Jatavallabhula, Krishna Murthy and Sen, Bipasha and Agarwal, Aditya and Rivera, Corban and Paul, William and Ellis, Kirsty and Chellappa, Rama and others},
  booktitle={2024 IEEE International Conference on Robotics and Automation (ICRA)},
  pages={5021--5028},
  year={2024},
  organization={IEEE}
}

@inproceedings{egtea2018,
  title={In the eye of beholder: Joint learning of gaze and actions in first person video},
  author={Li, Yin and Liu, Miao and Rehg, James M},
  booktitle={Proceedings of the European conference on computer vision (ECCV)},
  pages={639--655},
  year={2018}
}

@inproceedings{
pei2025egovideo,
title={Modeling Fine-Grained Hand-Object Dynamics for Egocentric Video Representation Learning},
author={Baoqi Pei and Yifei Huang and Jilan Xu and Guo Chen and Yuping He and Lijin Yang and Yali Wang and Weidi Xie and Yu Qiao and Fei Wu and Limin Wang},
booktitle={The Thirteenth International Conference on Learning Representations},
year={2025},
url={https://openreview.net/forum?id=P6G1Z6jkf3}
}

@INPROCEEDINGS{fun3Dsg,
  author={Zhang, Chenyangguang and Delitzas, Alexandros and Wang, Fangjinhua and Zhang, Ruida and Ji, Xiangyang and Pollefeys, Marc and Engelmann, Francis},
  booktitle={2025 IEEE/CVF Conference on Computer Vision and Pattern Recognition (CVPR)}, 
  title={Open-Vocabulary Functional 3D Scene Graphs for Real-World Indoor Spaces}, 
  year={2025},
  volume={},
  number={},
  pages={19401-19413},
  doi={10.1109/CVPR52734.2025.01807}}

@inproceedings{
wei2022chain,
title={Chain of Thought Prompting Elicits Reasoning in Large Language Models},
author={Jason Wei and Xuezhi Wang and Dale Schuurmans and Maarten Bosma and brian ichter and Fei Xia and Ed H. Chi and Quoc V Le and Denny Zhou},
booktitle={Advances in Neural Information Processing Systems},
editor={Alice H. Oh and Alekh Agarwal and Danielle Belgrave and Kyunghyun Cho},
year={2022},
url={https://openreview.net/forum?id=_VjQlMeSB_J}
}

@inproceedings{alayrac2022flamingo,
author = {Alayrac, Jean-Baptiste and Donahue, Jeff and Luc, Pauline and Miech, Antoine and Barr, Iain and Hasson, Yana and Lenc, Karel and Mensch, Arthur and Millicah, Katie and Reynolds, Malcolm and Ring, Roman and Rutherford, Eliza and Cabi, Serkan and Han, Tengda and Gong, Zhitao and Samangooei, Sina and Monteiro, Marianne and Menick, Jacob and Borgeaud, Sebastian and Brock, Andrew and Nematzadeh, Aida and Sharifzadeh, Sahand and Binkowski, Mikolaj and Barreira, Ricardo and Vinyals, Oriol and Zisserman, Andrew and Simonyan, Karen},
title = {Flamingo: a visual language model for few-shot learning},
year = {2022},
isbn = {9781713871088},
publisher = {Curran Associates Inc.},
address = {Red Hook, NY, USA},
booktitle = {Proceedings of the 36th International Conference on Neural Information Processing Systems},
articleno = {1723},
numpages = {21},
location = {New Orleans, LA, USA},
series = {NIPS '22}
}

@article{qwenteam2025qwen3vl,
  title={Qwen3-vl technical report},
  author={{Qwen Team}},
  journal={arXiv preprint arXiv:2511.21631},
  year={2025}
}

@misc{qwen3.5,
    title  = {{Qwen3.5}: Towards Native Multimodal Agents},
    author = {{Qwen Team}},
    year   = {2026},
    month  = {February},
    url    = {https://qwen.ai/blog?id=qwen3.5}
}

@inproceedings{zhao2023learning,
  title={Learning video representations from large language models},
  author={Zhao, Yue and Misra, Ishan and Kr{\"a}henb{\"u}hl, Philipp and Girdhar, Rohit},
  booktitle={Proceedings of the IEEE/CVF conference on computer vision and pattern recognition},
  pages={6586--6597},
  year={2023}
}

@inproceedings{qi2026llavaction,
  title={LLaVAction: evaluating and training multi-modal large language models for action understanding},
  author={Qi, Haozhe and Ye, Shaokai and Mathis, Alexander and Mathis, Mackenzie W},
  booktitle={International Conference on Learning Representations},
  pages={21859--21883},
  year={2026}
}

@inproceedings{liu2023llava,
author = {Liu, Haotian and Li, Chunyuan and Wu, Qingyang and Lee, Yong Jae},
title = {Visual instruction tuning},
year = {2023},
publisher = {Curran Associates Inc.},
address = {Red Hook, NY, USA},
booktitle = {Proceedings of the 37th International Conference on Neural Information Processing Systems},
articleno = {1516},
numpages = {25},
location = {New Orleans, LA, USA},
series = {NIPS '23}
}

@misc{meta2025llama4,
  title  = {The {Llama 4} Herd: The Beginning of a New Era of Natively Multimodal AI Innovation},
  author = {{Meta AI}},
  year   = {2025},
  month  = {April},
  howpublished = {\url{https://ai.meta.com/blog/llama-4-multimodal-intelligence/}}
}

@article{gemmateam2025gemma3,
  title={Gemma 3 Technical Report},
  author={{Gemma Team}},
  journal={arXiv preprint arXiv:2503.19786},
  year={2025}
}

@article{ma2025bridging,
  title={Bridging Vision Language Models and Symbolic Grounding for Video Question Answering},
  author={Ma, Haodi and Pathak, Vyom and Wang, Daisy Zhe},
  journal={arXiv preprint arXiv:2509.11862},
  year={2025}
}

@inproceedings{rodin2025easg,
  title={EASG-Bench: Video Q\&A Benchmark with Egocentric Action Scene Graphs},
  author={Rodin, Ivan and Wu, Tz-Ying and Min, Kyle and Sridhar, Sharath Nittur and Furnari, Antonino and Tripathi, Subarna and Farinella, Giovanni Maria},
  booktitle={Proceedings of the IEEE/CVF International Conference on Computer Vision Workshops},
  pages={2732--2737},
  year={2025}
}

@article{glmv,
  title={GLM-4.5V and GLM-4.1V-Thinking: Towards Versatile Multimodal Reasoning with Scalable Reinforcement Learning},
  author={{GLM-V Team}},
  journal={arXiv preprint arXiv:2507.01006},
  year={2025}
}

@article{kimiteam2025kimivl,
  title={Kimi-VL Technical Report},
  author={{Kimi Team}},
  journal={arXiv preprint arXiv:2504.07491},
  year={2025}
}

@article{lin2022egocentric,
  title={Egocentric video-language pretraining},
  author={Lin, Kevin Qinghong and Wang, Jinpeng and Soldan, Mattia and Wray, Michael and Yan, Rui and Xu, Eric Z and Gao, Difei and Tu, Rong-Cheng and Zhao, Wenzhe and Kong, Weijie and others},
  journal={Advances in Neural Information Processing Systems},
  volume={35},
  pages={7575--7586},
  year={2022}
}

@inproceedings{brown2020gpt3,
author = {Brown, Tom B. and Mann, Benjamin and Ryder, Nick and Subbiah, Melanie and Kaplan, Jared and Dhariwal, Prafulla and Neelakantan, Arvind and Shyam, Pranav and Sastry, Girish and Askell, Amanda and Agarwal, Sandhini and Herbert-Voss, Ariel and Krueger, Gretchen and Henighan, Tom and Child, Rewon and Ramesh, Aditya and Ziegler, Daniel M. and Wu, Jeffrey and Winter, Clemens and Hesse, Christopher and Chen, Mark and Sigler, Eric and Litwin, Mateusz and Gray, Scott and Chess, Benjamin and Clark, Jack and Berner, Christopher and McCandlish, Sam and Radford, Alec and Sutskever, Ilya and Amodei, Dario},
title = {Language models are few-shot learners},
year = {2020},
isbn = {9781713829546},
publisher = {Curran Associates Inc.},
address = {Red Hook, NY, USA},
booktitle = {Proceedings of the 34th International Conference on Neural Information Processing Systems},
articleno = {159},
numpages = {25},
location = {Vancouver, BC, Canada},
series = {NIPS '20}
}

@inproceedings{yang2022empirical,
  title={An Empirical Study of GPT-3 for Few-Shot Knowledge-Based VQA},
  author={Yang, Zhengyuan and Gan, Zhe and Wang, Jianfeng and Hu, Xiaowei and Lu, Yumao and Liu, Zicheng and Wang, Lijuan},
  booktitle={Proceedings of the AAAI Conference on Artificial Intelligence},
  pages={3081--3089},
  year={2022}
}

@inproceedings{zhang2024llovi,
    title = "A Simple {LLM} Framework for Long-Range Video Question-Answering",
    author = "Zhang, Ce  and
      Lu, Taixi  and
      Islam, Md Mohaiminul  and
      Wang, Ziyang  and
      Yu, Shoubin  and
      Bansal, Mohit  and
      Bertasius, Gedas",
    editor = "Al-Onaizan, Yaser  and
      Bansal, Mohit  and
      Chen, Yun-Nung",
    booktitle = "Proceedings of the 2024 Conference on Empirical Methods in Natural Language Processing",
    month = nov,
    year = "2024",
    address = "Miami, Florida, USA",
    publisher = "Association for Computational Linguistics",
    url = "https://aclanthology.org/2024.emnlp-main.1209/",
    doi = "10.18653/v1/2024.emnlp-main.1209",
    pages = "21715--21737",
}

@inproceedings{yi2018neural,
    author = {Yi, Kexin and Wu, Jiajun and Gan, Chuang and Torralba, Antonio and Kohli, Pushmeet and Tenenbaum, Joshua B.},
    title = {Neural-symbolic VQA: disentangling reasoning from vision and language understanding},
    year = {2018},
    publisher = {Curran Associates Inc.},
    address = {Red Hook, NY, USA},
    booktitle = {Proceedings of the 32nd International Conference on Neural Information Processing Systems},
    pages = {1039–1050},
    numpages = {12},
    location = {Montr\'{e}al, Canada},
    series = {NIPS'18}
}

@INPROCEEDINGS{johnson2015imagegraphs,
  author={Johnson, Justin and Krishna, Ranjay and Stark, Michael and Li, Li-Jia and Shamma, David A. and Bernstein, Michael S. and Fei-Fei, Li},
  booktitle={2015 IEEE Conference on Computer Vision and Pattern Recognition (CVPR)}, 
  title={Image retrieval using scene graphs}, 
  year={2015},
  volume={},
  number={},
  pages={3668-3678},
  doi={10.1109/CVPR.2015.7298990}}

@inproceedings{xu2017scenegraph,
  title={Scene graph generation by iterative message passing},
  author={Xu, Danfei and Zhu, Yuke and Choy, Christopher B and Fei-Fei, Li},
  booktitle={2017 IEEE Conference on Computer Vision and Pattern Recognition (CVPR)},
  pages={3097--3106},
  year={2017},
  organization={IEEE}
}

@inproceedings{yao2023tree,
 author = {Yao, Shunyu and Yu, Dian and Zhao, Jeffrey and Shafran, Izhak and Griffiths, Tom and Cao, Yuan and Narasimhan, Karthik},
 booktitle = {Advances in Neural Information Processing Systems},
 editor = {A. Oh and T. Naumann and A. Globerson and K. Saenko and M. Hardt and S. Levine},
 pages = {11809--11822},
 publisher = {Curran Associates, Inc.},
 title = {Tree of Thoughts: Deliberate Problem Solving with Large Language Models},
 url = {https://proceedings.neurips.cc/paper_files/paper/2023/file/271db9922b8d1f4dd7aaef84ed5ac703-Paper-Conference.pdf},
 volume = {36},
 year = {2023}
}

@inproceedings{wang2023nlgraph,
author = {Wang, Heng and Feng, Shangbin and He, Tianxing and Tan, Zhaoxuan and Han, Xiaochuang and Tsvetkov, Yulia},
title = {Can language models solve graph problems in natural language?},
year = {2023},
publisher = {Curran Associates Inc.},
address = {Red Hook, NY, USA},
booktitle = {Proceedings of the 37th International Conference on Neural Information Processing Systems},
articleno = {1345},
numpages = {22},
location = {New Orleans, LA, USA},
series = {NIPS '23}
}

@article{ek100,
    title = "Rescaling Egocentric Vision: Collection, Pipeline and Challenges for EPIC-KITCHENS-100",
    author = "Dima Damen and Hazel Doughty and Giovanni Maria Farinella and Antonino Furnari and Evangelos Kazakos and Jian Ma and Davide Moltisanti and Jonathan Munro and Toby Perrett and Will Price and Michael Wray",
    year = "2022",
    month = jan,
    day = "1",
    doi = "10.1007/s11263-021-01531-2",
    language = "English",
    volume = "130",
    pages = "33--55",
    journal = "International Journal of Computer Vision",
    issn = "1573-1405",
    publisher = "Springer",
    number = "1",
}

@inproceedings{grauman2022ego4d,
  title={Ego4d: Around the world in 3,000 hours of egocentric video},
  author={Grauman, Kristen and Westbury, Andrew and Byrne, Eugene and Chavis, Zachary and Furnari, Antonino and Girdhar, Rohit and Hamburger, Jackson and Jiang, Hao and Liu, Miao and Liu, Xingyu and others},
  booktitle={Proceedings of the IEEE/CVF conference on computer vision and pattern recognition},
  pages={18995--19012},
  year={2022}
}

@inproceedings{johnson2018image_gen_sg,
  title={Image generation from scene graphs},
  author={Johnson, Justin and Gupta, Agrim and Fei-Fei, Li},
  booktitle={Proceedings of the IEEE conference on computer vision and pattern recognition},
  pages={1219--1228},
  year={2018}
}

@inproceedings{sayplan2023,
  title={SayPlan: Grounding Large Language Models using 3D Scene Graphs for Scalable Robot Task Planning},
  author={Rana, Krishan and Haviland, Jesse and Garg, Sourav and Abou-Chakra, Jad and Reid, Ian and S{\"u}nderhauf, Niko},
  booktitle={Proceedings of the 7th Conference on Robot Learning (CoRL)},
  pages={23--72},
  year={2023},
  organization={Proceedings of Machine Learning Research}
}

@INPROCEEDINGS{werby2024hovsg, 
              AUTHOR    = {Abdelrhman Werby AND Chenguang Huang AND Martin Büchner AND Abhinav Valada AND Wolfram Burgard}, 
              TITLE     = {{Hierarchical Open-Vocabulary 3D Scene Graphs for Language-Grounded Robot Navigation}}, 
              BOOKTITLE = {Proceedings of Robotics: Science and Systems}, 
              YEAR      = {2024}, 
              ADDRESS   = {Delft, Netherlands}, 
              MONTH     = {July}, 
              DOI       = {10.15607/RSS.2024.XX.077} 
          }

@inproceedings{koch2024open3dsg,
  title={Open3dsg: Open-vocabulary 3d scene graphs from point clouds with queryable objects and open-set relationships},
  author={Koch, Sebastian and Vaskevicius, Narunas and Colosi, Mirco and Hermosilla, Pedro and Ropinski, Timo},
  booktitle={Proceedings of the IEEE/CVF Conference on Computer Vision and Pattern Recognition},
  pages={14183--14193},
  year={2024}
}

@article{rosinol2020dsg,
  title={3D Dynamic Scene Graphs: Actionable Spatial Perception with Places, Objects, and Humans},
  author={Rosinol, Antoni and Gupta, Arjun and Abate, Marcus and Shi, Jingnan and Carlone, Luca},
  journal={Robotics: Science and Systems XVI},
  year={2020},
  publisher={Robotics: Science and Systems Foundation}
}

@inproceedings{ji2020action_genome,
  title={Action genome: Actions as compositions of spatio-temporal scene graphs},
  author={Ji, Jingwei and Krishna, Ranjay and Fei-Fei, Li and Niebles, Juan Carlos},
  booktitle={Proceedings of the IEEE/CVF conference on computer vision and pattern recognition},
  pages={10236--10247},
  year={2020}
}

@inproceedings{li2023blip,
  title={Blip-2: Bootstrapping language-image pre-training with frozen image encoders and large language models},
  author={Li, Junnan and Li, Dongxu and Savarese, Silvio and Hoi, Steven},
  booktitle={International conference on machine learning},
  pages={19730--19742},
  year={2023},
  organization={PMLR}
}

@inproceedings{liu2024tempcompass,
  title={Tempcompass: Do video llms really understand videos?},
  author={Liu, Yuanxin and Li, Shicheng and Liu, Yi and Wang, Yuxiang and Ren, Shuhuai and Li, Lei and Chen, Sishuo and Sun, Xu and Hou, Lu},
  booktitle={Findings of the Association for Computational Linguistics: ACL 2024},
  pages={8731--8772},
  year={2024}
}

@inproceedings{huang2025building,
  title={Building a mind palace: Structuring environment-grounded semantic graphs for effective long video analysis with llms},
  author={Huang, Zeyi and Ji, Yuyang and Wang, Xiaofang and Mehta, Nikhil and Xiao, Tong and Lee, Donghyun and Vanvalkenburgh, Sigmund and Zha, Shengxin and Lai, Bolin and Yu, Licheng and others},
  booktitle={Proceedings of the IEEE/CVF conference on computer vision and pattern recognition},
  pages={24169--24179},
  year={2025}
}

@article{mangalam2023egoschema,
  title={Egoschema: A diagnostic benchmark for very long-form video language understanding},
  author={Mangalam, Karttikeya and Akshulakov, Raiymbek and Malik, Jitendra},
  journal={Advances in Neural Information Processing Systems},
  volume={36},
  pages={46212--46244},
  year={2023}
}

@inproceedings{
fatemi2024talk,
title={Talk like a Graph: Encoding Graphs for Large Language Models},
author={Bahare Fatemi and Jonathan Halcrow and Bryan Perozzi},
booktitle={The Twelfth International Conference on Learning Representations},
year={2024},
url={https://openreview.net/forum?id=IuXR1CCrSi}
}

@inproceedings{gupta2023visual,
  title={Visual programming: Compositional visual reasoning without training},
  author={Gupta, Tanmay and Kembhavi, Aniruddha},
  booktitle={Proceedings of the IEEE/CVF conference on computer vision and pattern recognition},
  pages={14953--14962},
  year={2023}
}

@inproceedings{jiang2024manyshot,
  title={Many-Shot In-Context Learning in Multimodal Foundation Models},
  author={Jiang, Yixing and Irvin, Jeremy Andrew and Wang, Ji Hun and Chaudhry, Muhammad Ahmed and Chen, Jonathan H and Ng, Andrew Y},
  booktitle={ICML 2024 Workshop on In-Context Learning},
  year={2024},
  url={https://openreview.net/forum?id=j2rKwWXdcz}
}

@inproceedings{zhou2024visualICL,
  title={Visual in-context learning for large vision-language models},
  author={Zhou, Yucheng and Li, Xiang and Wang, Qianning and Shen, Jianbing},
  booktitle={Findings of the Association for Computational Linguistics: ACL 2024},
  pages={15890--15902},
  year={2024}
}

@article{dai2025gpt4ego,
  author={Dai, Guangzhao and Shu, Xiangbo and Wu, Wenhao and Yan, Rui and Zhang, Jiachao},
  journal={IEEE Transactions on Multimedia}, 
  title={GPT4Ego: Unleashing the Potential of Pre-Trained Models for Zero-Shot Egocentric Action Recognition}, 
  year={2025},
  volume={27},
  number={},
  pages={401-413},
  doi={10.1109/TMM.2024.3521658}
}

@inproceedings{li2025vlmsurvey,
  author={Li, Zongxia and Wu, Xiyang and Du, Hongyang and Liu, Fuxiao and Nghiem, Huy and Shi, Guangyao},
  booktitle={2025 IEEE/CVF Conference on Computer Vision and Pattern Recognition Workshops (CVPRW)}, 
  title={A Survey of State of the Art Large Vision Language Models: Alignment, Benchmark, Evaluations and Challenges}, 
  year={2025},
  volume={},
  number={},
  pages={1578-1597},
  doi={10.1109/CVPRW67362.2025.00147}
}

@inproceedings{plizzari2025egotempo,
      title={Omnia de EgoTempo: Benchmarking Temporal Understanding of Multi-Modal LLMs in Egocentric Videos}, 
      author={Chiara Plizzari and Alessio Tonioni and Yongqin Xian and Achin Kulshrestha and Federico Tombari},
      booktitle={Proceedings of the IEEE/CVF Conference on Computer Vision and Pattern Recognition},
      year={2025},
}

@inproceedings{dong-etal-2024-contamination,
    title = "Generalization or Memorization: Data Contamination and Trustworthy Evaluation for Large Language Models",
    author = "Dong, Yihong  and
      Jiang, Xue  and
      Liu, Huanyu  and
      Jin, Zhi  and
      Gu, Bin  and
      Yang, Mengfei  and
      Li, Ge",
    editor = "Ku, Lun-Wei  and
      Martins, Andre  and
      Srikumar, Vivek",
    booktitle = "Findings of the Association for Computational Linguistics: ACL 2024",
    month = aug,
    year = "2024",
    address = "Bangkok, Thailand",
    publisher = "Association for Computational Linguistics",
    url = "https://aclanthology.org/2024.findings-acl.716/",
    doi = "10.18653/v1/2024.findings-acl.716",
    pages = "12039--12050",
}

@inproceedings{zeng2023socratic,
    title={Socratic Models: Composing Zero-Shot Multimodal Reasoning with Language},
    author={Andy Zeng and Maria Attarian and brian ichter and Krzysztof Marcin Choromanski and Adrian Wong and Stefan Welker and Federico Tombari and Aveek Purohit and Michael S Ryoo and Vikas Sindhwani and Johnny Lee and Vincent Vanhoucke and Pete Florence},
    booktitle={The Eleventh International Conference on Learning Representations },
    year={2023},
    url={https://openreview.net/forum?id=G2Q2Mh3avow}
}

@inproceedings{rodin2024easg,
  title={Action scene graphs for long-form understanding of egocentric videos},
  author={Rodin, Ivan and Furnari, Antonino and Min, Kyle and Tripathi, Subarna and Farinella, Giovanni Maria},
  booktitle={Proceedings of the IEEE/CVF Conference on Computer Vision and Pattern Recognition},
  pages={18622--18632},
  year={2024}
}

@article{taluzzi2025scenenet,
  title={From pixels to graphs: using scene and knowledge graphs for hd-epic vqa challenge},
  author={Taluzzi, Agnese and Gesualdi, Davide and Santambrogio, Riccardo and Plizzari, Chiara and Palermo, Francesca and Mentasti, Simone and Matteucci, Matteo},
  journal={arXiv preprint arXiv:2506.08553},
  year={2025}
}

@inproceedings{yu2024eilev,
    title = "Eliciting In-Context Learning in Vision-Language Models for Videos Through Curated Data Distributional Properties",
    author = "Yu, Keunwoo Peter  and
      Zhang, Zheyuan  and
      Hu, Fengyuan  and
      Storks, Shane  and
      Chai, Joyce",
    editor = "Al-Onaizan, Yaser  and
      Bansal, Mohit  and
      Chen, Yun-Nung",
    booktitle = "Proceedings of the 2024 Conference on Empirical Methods in Natural Language Processing",
    month = nov,
    year = "2024",
    address = "Miami, Florida, USA",
    publisher = "Association for Computational Linguistics",
    url = "https://aclanthology.org/2024.emnlp-main.1137/",
    doi = "10.18653/v1/2024.emnlp-main.1137",
    pages = "20416--20431"
}

@article{chen2023manipulating,
  title={Manipulating the label space for in-context classification},
  author={Chen, Haokun and Yang, Xu and Huang, Yuhang and Wu, Zihan and Wang, Jing and Geng, Xin},
  journal={arXiv preprint arXiv:2312.00351},
  year={2023}
}

@inproceedings{zong2025vlicl,
title={{VL}-{ICL} Bench: The Devil in the Details of Multimodal In-Context Learning},
author={Yongshuo Zong and Ondrej Bohdal and Timothy Hospedales},
booktitle={The Thirteenth International Conference on Learning Representations},
year={2025},
url={https://openreview.net/forum?id=cpGPPLLYYx}
}

@inproceedings{kirillov2023sam,
  title={Segment anything},
  author={Kirillov, Alexander and Mintun, Eric and Ravi, Nikhila and Mao, Hanzi and Rolland, Chloe and Gustafson, Laura and Xiao, Tete and Whitehead, Spencer and Berg, Alexander C and Lo, Wan-Yen and others},
  booktitle={Proceedings of the IEEE/CVF international conference on computer vision},
  pages={4015--4026},
  year={2023}
}

@inproceedings{liu2024groundingdino,
  title={Grounding dino: Marrying dino with grounded pre-training for open-set object detection},
  author={Liu, Shilong and Zeng, Zhaoyang and Ren, Tianhe and Li, Feng and Zhang, Hao and Yang, Jie and Jiang, Qing and Li, Chunyuan and Yang, Jianwei and Su, Hang and others},
  booktitle={European conference on computer vision},
  pages={38--55},
  year={2024},
  organization={Springer}
}

@inproceedings{hd-epic,
title = "HD-EPIC: A Highly-Detailed Egocentric Video Dataset",
author = "Toby Perrett and Ahmad Darkhalil and Saptarshi Sinha and Omar Emara and Sam Pollard and Kranti Parida and Kaiting Liu and Prajwal Gatti and Siddhant Bansal and Kevin Flanagan and Jacob Chalk and Zhifan Zhu and Rhodri Guerrier and Fahd Abdelazim and Bin Zhu and Davide Moltisanti and Michael Wray and Hazel Doughty and Dima Damen",
year = "2025",
doi = "10.1109/CVPR52734.2025.02226",
isbn = "9798331543655",
publisher = "IEEE",
pages = "23901--23913",
booktitle = "2025 IEEE/CVF Conference on Computer Vision and Pattern Recognition (CVPR)",
}

@inproceedings{vllm,
author = {Kwon, Woosuk and Li, Zhuohan and Zhuang, Siyuan and Sheng, Ying and Zheng, Lianmin and Yu, Cody Hao and Gonzalez, Joseph and Zhang, Hao and Stoica, Ion},
title = {Efficient Memory Management for Large Language Model Serving with PagedAttention},
year = {2023},
isbn = {9798400702297},
publisher = {Association for Computing Machinery},
address = {New York, NY, USA},
url = {https://doi.org/10.1145/3600006.3613165},
doi = {10.1145/3600006.3613165},
booktitle = {Proceedings of the 29th Symposium on Operating Systems Principles},
pages = {611–626},
numpages = {16},
location = {Koblenz, Germany},
series = {SOSP '23}
}
}

\clearpage
\appendix
\maketitlesupplementary



\section{Model details}
\label{app:models}

The selection of models used in this paper (see \cref{tab:vlms}) aims for breadth along two axes: model family, so that conclusions about the semantic bottleneck are not specific to a single training recipe, and parameter scale, which lets us check whether the bottleneck eases or persists as capacity grows. The largest checkpoints are quantized only as far as needed to fit our inference budget, so accuracy differences across rows reflect representational choices rather than precision loss.

\begin{table*}[t]
\centering
\caption{\textbf{Vision-Language Models evaluated.} Eleven instruction-tuned VLMs from six open-weight families, grouped into three size regimes. MoE rows report \emph{total / active} parameters; dense rows report total. Context lengths are the native pre-training windows reported by the model authors. Models marked $\ast$ run from quantized checkpoints to fit inference within four L40S GPUs. All models are evaluated with reasoning modes disabled and greedy decoding ($T\!=\!0$).}
\label{tab:vlms}
\small
\setlength{\tabcolsep}{6pt}
\begin{tabular}{lcccl}
\toprule
Model & Params (Total / Active) & Type & Context Length & Quantization \\
\midrule
\multicolumn{5}{l}{\textit{Small ($<20$B)}} \\
Qwen3.5-2B \cite{qwen3.5}                          & 2B           & Dense & 256K & - \\
Gemma-3-4B-it \cite{gemmateam2025gemma3}                      & 4B           & Dense & 128K &  - \\
GLM-4.6V-Flash \cite{glmv}                     & 9B           & Dense & 128K & - \\
Kimi-VL-A3B-Instruct \cite{kimiteam2025kimivl}               & 16B / 3B     & MoE   & 128K & - \\
\midrule
\multicolumn{5}{l}{\textit{Medium ($20$--$100$B)}} \\
Gemma-3-27B-it \cite{gemmateam2025gemma3}                    & 27B          & Dense & 128K & - \\
Qwen3-VL-32B-Instruct \cite{qwenteam2025qwen3vl}              & 32B          & Dense & 256K & - \\
Qwen3.5-35B-A3B   \cite{qwen3.5}                  & 35B / 3B     & MoE   & 256K & - \\
\midrule
\multicolumn{5}{l}{\textit{Large ($>100$B), quantized}} \\
GLM-4.6V$^{\ast}$  \cite{glmv}                 & 106B / 12B         & MoE & 128K & FP8 \\
Llama-4-Scout-17B-16E$^{\ast}$   \cite{meta2025llama4}    & 109B / 17B   & MoE   & 10M & FP8 \\
Qwen3.5-122B-A10B$^{\ast}$ \cite{qwen3.5}         & 122B / 10B   & MoE   & 256K & FP8  \\
Qwen3-VL-235B-A22B-Instruct$^{\ast}$ \cite{qwenteam2025qwen3vl} & 235B / 22B  & MoE   & 256K & AWQ-Int4 \\
\bottomrule
\end{tabular}
\end{table*}

\section{Prompt templates}
\label{app:prompts}

\subsection{Stage-1 description prompt}
\label{app:stage1_prompt}

For each temporal window $W_t$, we prompt the VLM to generate a linguistic description of the observed interaction. To ensure the reliability of the "image-to-symbolic" transition, we include a self-correction instruction within the prompt. The exact template is provided below:

\begin{tcolorbox}[
    breakable,
    colback=gray!5, 
    colframe=gray!50, 
    arc=1mm, 
    boxrule=0.5pt, 
    title=Prompt Template: Stage-1 (Action Description),
    fonttitle=\bfseries\small,
    fontupper=\small\ttfamily
]
Describe the main action the hands or the camera wearer are performing in the egocentric video frames. If visible, provide information about the right and left hands.

Once you have the description, double-check---based on the visual details in the frames---the object(s) involved in the action to ensure the object name is correct and not confused with a visually similar object.
\end{tcolorbox}

\subsection{Stage-2 graph summarization prompt}
\label{app:stage2_prompt}

In the second stage, the free-text descriptions generated in Stage-1 are mapped onto a structured symbolic space. We prompt the VLM to act as a deterministic parser, converting natural language into a JSON-formatted graph representation. This stage enforces a strict schema to facilitate the subsequent temporal graph aggregation. The prompt is structured as follows:

\begin{tcolorbox}[
    breakable,
    colback=gray!5, 
    colframe=gray!50, 
    arc=1mm, 
    boxrule=0.5pt, 
    title=Prompt Template: Stage-2 (Graph Summarization),
    fonttitle=\bfseries\small,
    fontupper=\small\ttfamily
]
You are specialized on creating graphs in JSON format from a description. The graph should summarize the main action described.

Here the graph schema:
\begin{center}
\begin{minipage}{0.95\linewidth}
\begin{footnotesize}
\begin{verbatim}
{
 "relations": [
  {
   "visibility": "hand_left"
                | "hand_right"
                | "hand_both"
                | "camera_wearer",
   "object": {
    "label": "name of the direct
              object (1 word)",
    "attribute": "most descriptive
                  attribute (1 word)"
   },
   "interaction": {
    "type": "main action verb
             (1 word in continuous form)"
   }
  }
 ],
}
\end{verbatim}
\end{footnotesize}
\end{minipage}
\end{center}

Use "hand\_both"  only when left and right hands physically cooperate on the same object with identical interaction. \\
If hands are in the description, do not include camera wearer information in the graph. \\
If there are descriptions from different frames, include in the graph the information of the last frame. \\
Include only one relation per same visibility entity. \\
\end{tcolorbox}

\subsection{Action recognition prompts}
\label{app:action_prompt}

We detail the prompt templates used for closed-set action recognition across the three evaluation settings considered in this work:
\emph{frames only}, \emph{graphs only}, and \emph{graphs + ICL}. All three
settings query the same VLM with the same user-side template; they differ
only in (i) the input artifact attached to the user message and (ii) whether
the system prompt includes labeled demonstrations. This isolates the effect
of the input representation from prompt-specific tuning.
\paragraph{Placeholders.} Throughout the templates, fields of the form
\texttt{<NAME>} are instantiated per setting, per dataset, or per example:
\begin{itemize}
  \item \texttt{<INPUT\_MODALITY>} -- the input representation:
        \emph{video sequences} (frames-only setting) or
        \emph{temporal dynamic graphs} (graph-based settings).
  \item \texttt{<MODALITY\_SPECIFICATION>} -- the concrete artifact passed
        to the model: \emph{sequence of frames sampled from a video} or
        \emph{temporal graph representation}.
  \item \texttt{<ACTION\_FORMAT>} -- the dataset-dependent label format:
        \texttt{"action\_id"} for EGTEA and \texttt{"verb:noun"} for
        Epic-Kitchens-100.
  \item \texttt{<ACTIONS\_OPTIONS>} -- the closed-set vocabulary of
        candidate action labels for the target dataset, formatted according
        to \texttt{<ACTION\_FORMAT>}.
  \item \texttt{<ICL\_EXAMPLES>} -- a sequence of $k$ labeled $(\mathrm{graph}, \mathrm{action})$ demonstrations per candidate class, drawn from the
        training split. Used only in the graphs + ICL setting.
  \item \texttt{<RESPONSE>} -- the required output schema: a JSON object
        containing the top-5 predicted labels (in \texttt{<ACTION\_FORMAT>})
        ranked by the model's self-reported confidence.
  \item \texttt{<EXAMPLE>} -- one illustrative JSON instance conforming to
        \texttt{<RESPONSE>}.
  \item \texttt{<INVALID>}, \texttt{<ERROR\_MSG>} -- the model's previous
        malformed output and the corresponding parser diagnostic; used only
        by the correction prompt.
\end{itemize}

\paragraph{System prompt (frames-only and graphs-only).}
The non-ICL system prompt is shared between the frames-only and graphs-only
conditions, with the input modality templated:

\begin{tcolorbox}[
    breakable,
    colback=gray!5, colframe=gray!50, arc=1mm, boxrule=0.5pt,
    title=Prompt Template: System,
    fonttitle=\bfseries\small,
    fontupper=\small\ttfamily
]
You are an expert at analyzing <INPUT\_MODALITY> and identifying actions.

Given a <MODALITY\_SPECIFICATION>, identify which action is most likely being
performed by the hands or the camera wearer.

Return your answer in JSON format with <RESPONSE>.
\end{tcolorbox}

\paragraph{System prompt (graphs + ICL).}
For the graph-based ICL setting, the system prompt is specialized to graphs
and prepends $k$ labeled demonstrations per candidate class from the training set,  so that the model can ground its interpretation of node and edge semantics in concrete examples:

\begin{tcolorbox}[
    breakable,
    colback=gray!5, colframe=gray!50, arc=1mm, boxrule=0.5pt,
    title=Prompt Template: System (ICL),
    fonttitle=\bfseries\small,
    fontupper=\small\ttfamily
]
You are an expert at analyzing temporal dynamic graphs and identifying actions.

Below are labeled examples of temporal graphs and their corresponding actions.
Use these as reference to understand the graph structure and action patterns:

<ICL\_EXAMPLES>

Now, given the following temporal graph, identify which action is most likely
being performed by the hands or the camera wearer.

Return your answer in JSON format with <RESPONSE>.
\end{tcolorbox}

\paragraph{User prompt.}
The user-side prompt differs slightly between the frames-only and the
graph-based settings, reflecting the different nature of the input. The
final block (candidate list, response schema, and example) is identical
across all three settings; only the lead-in and the input modality vary.

\begin{tcolorbox}[
    breakable,
    colback=gray!5, colframe=gray!50, arc=1mm, boxrule=0.5pt,
    title=Prompt Template: User (frames-only),
    fonttitle=\bfseries\small,
    fontupper=\small\ttfamily
]
Analyze these 16 frames from a video sequence and determine which action is
being performed.

Choose the top 5 most likely actions from this list:
<ACTIONS\_OPTIONS>

Return your answer in JSON format with <RESPONSE>.

Example: <EXAMPLE>
\end{tcolorbox}

\noindent The 16 sampled frames are attached as image content blocks
immediately before this text. For the graph-based settings (graphs only and
graphs + ICL), the input is provided as text inside the user message:

\begin{tcolorbox}[
    breakable,
    colback=gray!5, colframe=gray!50, arc=1mm, boxrule=0.5pt,
    title=Prompt Template: User (graph-based settings),
    fonttitle=\bfseries\small,
    fontupper=\small\ttfamily
]
Analyze the following temporal dynamic graph extracted from a video sequence
and determine which action is being performed.

GRAPH DATA:
<GRAPH\_CONTENT>

Choose the top 5 most likely actions from this list:
<ACTIONS\_OPTIONS>

Return your answer in JSON format with <RESPONSE>.

Example: <EXAMPLE>
\end{tcolorbox}

\paragraph{Correction prompt.}
VLM outputs that violate the response schema (wrong number of
predictions, labels outside the closed set, duplicate predictions, or
malformed JSON) are passed through a repair step using the template
below. The correction loop is invoked iteratively for up to two
repetitions. To keep the correction context manageable for datasets
with large label spaces, the candidate list shown in the correction
prompt is truncated to at most $50$ entries.

\begin{tcolorbox}[
    breakable,
    colback=gray!5, colframe=gray!50, arc=1mm, boxrule=0.5pt,
    title=Prompt Template: Correction,
    fonttitle=\bfseries\small,
    fontupper=\small\ttfamily
]
Your previous answer was invalid. You provided:

<INVALID>

Reason why it is invalid: <ERROR\_MSG>

Fix it using these rules:

- exactly 5 predictions

- each action must be chosen from <ACTIONS\_OPTIONS>

- all 5 <ACTION\_FORMAT> must be distinct

- return JSON only
\end{tcolorbox}

\section{Per-model results}
\label{app:supp_per_model}

\Cref{tab:full_results_all_models,tab:ek100_top1_top5_vna} 
report the per-model results underlying the aggregate trends in the main paper, for EGTEA and EK100 respectively. The full EGTEA table covers all eleven backbones; the EK100 sweep is restricted to a representative set of three models for compute reasons (GLM-4.6V-Flash, Kimi-VL-A3B, GLM-4.6V$^{*}$) and additionally varies the pool coverage and number of in-context examples per class. Two observations are worth noting. First, on EGTEA the \emph{Graphs only} $\to$ \emph{Graphs + ICL} progression holds across nearly every backbone, with the clear exception of Gemma-3 (both 4B and 27B), where graphs alone are the strongest modality overall and demonstrations degrade Top-1/3/5 accuracy. Second, the optimal number of in-context examples is model-dependent on EGTEA: most models prefer $k\!=\!2$, while Qwen3-VL-235B$^{*}$ peaks at $k\!=\!1$ and Qwen3-VL-32B and Llama-4-Scout$^{*}$ are flat between the two. On EK100 the picture is consistent with the pool-coverage sweep in the main paper: $k\!=\!1$ with a $100$-class pool is competitive across all three models.

\begin{table*}
  \caption{\textbf{EGTEA action recognition} across models and input modalities. We report MCA, Top-1, Top-3, and Top-5 accuracy. Within each model block, the best entry per metric is shown in \best{bold}, and the second-best is \underline{underlined}; the modality with the highest average across the four metrics is \colorbox{lightgray}{highlighted}. ICL rows (\textit{Graphs ICL $k$ ex}) report mean $\pm$ SEM over 5 demonstration draws. The \textit{Overall} block reports the macro-average across models for each input modality.}
  \label{tab:full_results_all_models}
  \centering
  \small
  \begin{tabular}{llcccc}
    \toprule
    Model & Input Modality & MCA & Top-1 & Top-3 & Top-5 \\
    \midrule

    \multirow{4}{*}{Qwen3.5-2B}
      & Frames only & \third{25.65} & 32.84 & \third{49.60} & \best{56.53} \\
      & Graphs only & 20.55 & \third{34.97} & 48.27 & 52.08 \\
      & Graphs ICL 1 ex & \second{30.90 $\pm$ 0.30} & \best{38.70 $\pm$ 0.32} & \best{51.64 $\pm$ 0.61} & \third{55.94 $\pm$ 0.67} \\
      & \bestoverall{Graphs ICL 2 ex} & \bestoverall{\best{32.24 $\pm$ 0.15}} & \bestoverall{\second{38.15 $\pm$ 0.21}} & \bestoverall{\second{51.55 $\pm$ 0.26}} & \bestoverall{\second{56.39 $\pm$ 0.36}} \\
    \midrule

    \multirow{4}{*}{Gemma-3-4B}
      & Frames only & 8.72 & 15.03 & 27.60 & 34.22 \\
      & \bestoverall{Graphs only} & \bestoverall{\best{14.24}} & \bestoverall{\best{32.29}} & \bestoverall{\best{43.72}} & \bestoverall{\best{50.74}} \\
      & Graphs ICL 1 ex & \second{13.50 $\pm$ 0.21} & \second{27.81 $\pm$ 0.26} & \second{36.58 $\pm$ 0.32} & \second{40.08 $\pm$ 0.49} \\
      & Graphs ICL 2 ex & \third{12.05 $\pm$ 0.18} & \third{23.80 $\pm$ 0.37} & \third{33.03 $\pm$ 0.14} & \third{37.02 $\pm$ 0.10} \\
    \midrule

    \multirow{4}{*}{GLM-4.6V-Flash}
      & Frames only & 27.11 & 39.66 & 51.04 & 54.70 \\
      & Graphs only & \third{31.60} & \third{44.96} & \second{61.82} & \best{67.06} \\
      & Graphs ICL 1 ex & \second{33.30 $\pm$ 0.32} & \second{46.40 $\pm$ 0.23} & \third{60.97 $\pm$ 0.25} & \third{64.81 $\pm$ 0.28} \\
      & \bestoverall{Graphs ICL 2 ex} & \bestoverall{\best{36.58 $\pm$ 0.38}} & \bestoverall{\best{46.82 $\pm$ 0.26}} & \bestoverall{\best{62.02 $\pm$ 0.29}} & \bestoverall{\second{65.55 $\pm$ 0.35}} \\
    \midrule

    \multirow{4}{*}{Kimi-VL-A3B}
      & Frames only & 20.03 & 34.03 & 45.75 & 50.15 \\
      & Graphs only & \third{20.97} & \third{37.88} & \third{51.34} & \third{54.25} \\
      & Graphs ICL 1 ex & \second{27.45 $\pm$ 0.50} & \second{40.35 $\pm$ 0.28} & \second{52.61 $\pm$ 0.27} & \second{55.53 $\pm$ 0.36} \\
      & \bestoverall{Graphs ICL 2 ex} & \bestoverall{\best{29.03 $\pm$ 0.50}} & \bestoverall{\best{41.37 $\pm$ 0.35}} & \bestoverall{\best{54.26 $\pm$ 0.51}} & \bestoverall{\best{57.14 $\pm$ 0.40}} \\
    \midrule

    \multirow{4}{*}{Gemma-3-27B}
      & Frames only & 23.08 & 34.72 & 51.04 & 57.67 \\
      & \bestoverall{Graphs only} & \bestoverall{\second{26.77}} & \bestoverall{\best{40.60}} & \bestoverall{\best{59.10}} & \bestoverall{\best{67.51}} \\
      & Graphs ICL 1 ex & \best{26.95 $\pm$ 0.60} & \second{39.13 $\pm$ 0.42} & \third{56.29 $\pm$ 0.24} & \third{64.27 $\pm$ 0.46} \\
      & Graphs ICL 2 ex & \third{24.90 $\pm$ 0.43} & \third{38.35 $\pm$ 0.54} & \second{56.54 $\pm$ 0.24} & \second{64.90 $\pm$ 0.08} \\
    \midrule

    \multirow{4}{*}{Qwen3-VL-32B}
      & Frames only & 34.62 & 47.58 & 65.43 & 72.50 \\
      & Graphs only & \third{37.18} & \third{50.20} & \third{69.93} & \third{77.65} \\
      & \bestoverall{Graphs ICL 1 ex} & \bestoverall{\best{42.53 $\pm$ 0.52}} & \bestoverall{\second{53.18 $\pm$ 0.47}} & \bestoverall{\best{72.22 $\pm$ 0.41}} & \bestoverall{\best{78.35 $\pm$ 0.17}} \\
      & Graphs ICL 2 ex & \second{42.38 $\pm$ 0.38} & \best{53.42 $\pm$ 0.22} & \second{71.50 $\pm$ 0.11} & \second{77.78 $\pm$ 0.30} \\
    \midrule

    \multirow{4}{*}{Qwen3.5-35B-A3B}
      & Frames only & \third{50.53} & \best{57.27} & \third{74.73} & 80.02 \\
      & Graphs only & 45.34 & 52.87 & 73.99 & \third{80.27} \\
      & \bestoverall{Graphs ICL 1 ex} & \bestoverall{\second{51.77 $\pm$ 0.27}} & \bestoverall{\second{55.87 $\pm$ 0.28}} & \bestoverall{\best{75.50 $\pm$ 0.35}} & \bestoverall{\best{82.97 $\pm$ 0.20}} \\
      & Graphs ICL 2 ex & \best{53.10 $\pm$ 0.54} & \third{53.97 $\pm$ 0.35} & \second{75.21 $\pm$ 0.37} & \second{82.74 $\pm$ 0.12} \\
    \midrule

    \multirow{4}{*}{GLM-4.6V*}
      & Frames only & 35.84 & 48.47 & 66.62 & 72.70 \\
      & Graphs only & \third{42.37} & \third{52.03} & \second{70.82} & \second{78.54} \\
      & Graphs ICL 1 ex & \second{45.68 $\pm$ 0.38} & \second{53.04 $\pm$ 0.34} & \third{70.60 $\pm$ 0.19} & \third{75.82 $\pm$ 0.17} \\
      & \bestoverall{Graphs ICL 2 ex} & \bestoverall{\best{47.68 $\pm$ 0.30}} & \bestoverall{\best{54.53 $\pm$ 0.21}} & \bestoverall{\best{72.64 $\pm$ 0.23}} & \bestoverall{\best{79.06 $\pm$ 0.33}} \\
    \midrule

    \multirow{4}{*}{Llama-4-Scout*}
      & Frames only & 17.38 & 30.96 & 45.90 & 50.99 \\
      & Graphs only & \third{24.97} & \third{40.21} & \third{61.67} & \best{70.28} \\
      & \bestoverall{Graphs ICL 1 ex} & \bestoverall{\best{30.01 $\pm$ 0.36}} & \bestoverall{\best{44.61 $\pm$ 0.39}} & \bestoverall{\second{62.75 $\pm$ 0.36}} & \bestoverall{\third{69.59 $\pm$ 0.43}} \\
      & Graphs ICL 2 ex & \second{29.68 $\pm$ 0.27} & \second{44.35 $\pm$ 0.13} & \best{62.81 $\pm$ 0.25} & \second{69.72 $\pm$ 0.34} \\
    \midrule

    \multirow{4}{*}{Qwen3.5-122B*}
      & Frames only & \third{51.73} & \best{58.06} & 74.83 & 80.81 \\
      & Graphs only & 47.17 & 53.61 & \third{76.16} & \third{83.93} \\
      & Graphs ICL 1 ex & \second{54.47 $\pm$ 0.48} & \second{57.74 $\pm$ 0.38} & \best{78.96 $\pm$ 0.23} & \best{85.55 $\pm$ 0.23} \\
      & \bestoverall{Graphs ICL 2 ex} & \bestoverall{\best{56.08 $\pm$ 0.54}} & \bestoverall{\third{57.59 $\pm$ 0.43}} & \bestoverall{\second{78.25 $\pm$ 0.31}} & \bestoverall{\second{85.10 $\pm$ 0.34}} \\
    \midrule

    \multirow{4}{*}{Qwen3-VL-235B*}
      & Frames only & 39.21 & 49.65 & 69.83 & 78.59 \\
      & Graphs only & \third{40.56} & \third{52.32} & \third{75.07} & \third{83.04} \\
      & Graphs ICL 1 ex & \best{47.02 $\pm$ 0.41} & \best{56.36 $\pm$ 0.38} & \second{78.13 $\pm$ 0.27} & \second{85.52 $\pm$ 0.21} \\
      & \bestoverall{Graphs ICL 2 ex} & \bestoverall{\second{46.07 $\pm$ 0.51}} & \bestoverall{\second{56.20 $\pm$ 0.34}} & \bestoverall{\best{79.15 $\pm$ 0.38}} & \bestoverall{\best{86.79 $\pm$ 0.21}} \\
    \midrule

    \multirow{4}{*}{Overall}
      & Frames only & 30.35 & 40.75 & 56.58 & 62.63 \\
      & Graphs only & \third{31.97} & \third{44.72} & \third{62.90} & \best{69.58} \\
      & {Graphs ICL 1 ex} & {\second{36.69}} & {\best{46.66}} & {\second{63.30}} & {\third{68.95}} \\
      & \bestoverall{Graphs ICL 2 ex} & \bestoverall{\best{37.25}} & \bestoverall{\second{46.23}} & \bestoverall{\best{63.36}} & \bestoverall{\second{69.29}} \\
    \bottomrule
  \end{tabular}
\end{table*}

\begin{table*}
  \caption{\textbf{Epic-Kitchens-100 action recognition} across models and input modalities. We report Top-1 and Top-5 accuracy for verb, noun, and action prediction; legend conventions follow \cref{tab:full_results_all_models}. ICL rows sweep the number of in-context examples per class ($k\!\in\!\{1,2\}$) and the number of classes the demonstration pool covers (the 50 or 100 most frequent), and report mean $\pm$ SEM over 5 demonstration draws. The \textit{Overall} block macro-averages each modality across models.}
  \label{tab:ek100_top1_top5_vna}
  \centering
  \small
  \setlength{\tabcolsep}{5pt}
  \resizebox{\textwidth}{!}{
  \begin{tabular}{llcccccc}
    \toprule
    \multirow{2}{*}{Model} & \multirow{2}{*}{Input Modality}
    & \multicolumn{3}{c}{Top-1 Acc.}
    & \multicolumn{3}{c}{Top-5 Acc.} \\
    \cmidrule(lr){3-5} \cmidrule(lr){6-8}
    & & Verb & Noun & Action & Verb & Noun & Action \\
    \midrule

    \multirow{6}{*}{GLM-4.6V-Flash}
      & Frames only                  & 24.54 & 23.71 & 9.91 & 26.53 & 35.44 & 13.83 \\
      & Graphs only                  & 18.76 & 24.09 & 6.68 & 26.64 & 38.49 & 12.86 \\
      & Graphs ICL (1 ex / 50 cls)   & 28.38 $\pm$ 0.64 & 26.24 $\pm$ 0.07 & 10.41 $\pm$ 0.13 & 39.81 $\pm$ 1.01 & 40.29 $\pm$ 0.13 & 19.74 $\pm$ 0.37 \\
      & \bestoverall{Graphs ICL (2 ex / 50 cls)} & \bestoverall{\second{28.55 $\pm$ 0.34}} & \bestoverall{26.97 $\pm$ 0.10} & \bestoverall{\best{10.66 $\pm$ 0.09}} & \bestoverall{\best{40.92 $\pm$ 0.67}} & \bestoverall{\best{41.42 $\pm$ 0.14}} & \bestoverall{\best{20.88 $\pm$ 0.32}} \\
      & Graphs ICL (1 ex / 100 cls)  & \best{28.59 $\pm$ 0.40} & \second{27.22 $\pm$ 0.12} & \second{10.65 $\pm$ 0.10} & \second{40.42 $\pm$ 0.65} & \second{40.95 $\pm$ 0.28} & \second{20.10 $\pm$ 0.22} \\
      & Graphs ICL (2 ex / 100 cls)  & 28.20 $\pm$ 0.36 & \best{27.45 $\pm$ 0.15} & 10.55 $\pm$ 0.12 & 37.60 $\pm$ 1.71 & 40.73 $\pm$ 1.17 & 20.06 $\pm$ 0.33 \\
    \midrule

    \multirow{6}{*}{Kimi-VL-A3B}
      & Frames only                  & \best{32.09} & 17.67 & 6.33 & \best{38.29} & 28.40 & 12.48 \\
      & Graphs only                  & 25.77 & 22.04 & 7.26 & 29.34 & \best{36.05} & 12.88 \\
      & Graphs ICL (1 ex / 50 cls)   & 26.87 $\pm$ 0.14 & \second{23.97 $\pm$ 0.11} & 8.52 $\pm$ 0.05 & 33.16 $\pm$ 0.26 & \second{34.22 $\pm$ 0.16} & \second{13.75 $\pm$ 0.17} \\
      & Graphs ICL (2 ex / 50 cls)   & \second{27.30 $\pm$ 0.10} & 23.83 $\pm$ 0.17 & \second{8.85 $\pm$ 0.10} & 32.91 $\pm$ 0.32 & 32.99 $\pm$ 0.10 & 13.61 $\pm$ 0.12 \\
      & \bestoverall{Graphs ICL (1 ex / 100 cls)} & \bestoverall{\best{27.59 $\pm$ 0.33}} & \bestoverall{\best{24.00 $\pm$ 0.08}} & \bestoverall{\best{9.09 $\pm$ 0.15}} & \bestoverall{33.30 $\pm$ 0.14} & \bestoverall{33.58 $\pm$ 0.17} & \bestoverall{\best{13.93 $\pm$ 0.11}} \\
      & Graphs ICL (2 ex / 100 cls)  & 26.86 $\pm$ 0.08 & 23.85 $\pm$ 0.12 & 8.76 $\pm$ 0.04 & 31.09 $\pm$ 0.30 & 32.43 $\pm$ 0.17 & 12.94 $\pm$ 0.04 \\
    \midrule

    \multirow{6}{*}{GLM-4.6V*}
      & Frames only                  & 27.18 & 26.22 & 10.60 & 32.73 & 43.12 & 17.68 \\
      & Graphs only                  & 24.88 & 25.81 & 9.01 & 34.43 & 41.93 & 16.32 \\
      & Graphs ICL (1 ex / 50 cls)   & 33.45 $\pm$ 0.20 & 28.14 $\pm$ 0.17 & 13.32 $\pm$ 0.11 & 46.45 $\pm$ 0.33 & 44.88 $\pm$ 0.15 & 25.31 $\pm$ 0.12 \\
      & Graphs ICL (2 ex / 50 cls)   & 34.48 $\pm$ 0.15 & 28.24 $\pm$ 0.13 & 13.95 $\pm$ 0.10 & 47.21 $\pm$ 0.58 & 45.77 $\pm$ 0.55 & 26.57 $\pm$ 0.11 \\
      & \bestoverall{Graphs ICL (1 ex / 100 cls)} & \bestoverall{\second{34.58 $\pm$ 0.21}} & \bestoverall{\best{29.57 $\pm$ 0.13}} & \bestoverall{\second{14.02 $\pm$ 0.15}} & \bestoverall{\best{47.96 $\pm$ 0.29}} & \bestoverall{\best{46.66 $\pm$ 0.05}} & \bestoverall{\second{26.69 $\pm$ 0.15}} \\
      & Graphs ICL (2 ex / 100 cls)  & \best{34.64 $\pm$ 0.05} & \second{29.39 $\pm$ 0.16} & \best{14.14 $\pm$ 0.08} & \second{47.68 $\pm$ 0.14} & \second{46.45 $\pm$ 0.11} & \best{26.88 $\pm$ 0.08} \\
    \midrule

    \multirow{6}{*}{Overall}
      & Frames only                  & 27.94 & 22.53 & 8.95 & 32.52 & 35.65 & 14.66 \\
      & Graphs only                  & 23.14 & 23.98 & 7.65 & 30.14 & 38.82 & 14.02 \\
      & Graphs ICL (1 ex / 50 cls)   & 29.57 & 26.12 & 10.75 & 39.81 & 39.80 & 19.60 \\
      & Graphs ICL (2 ex / 50 cls)   & \second{30.11} & 26.35 & \second{11.15} & \second{40.35} & \second{40.06} & \best{20.35} \\
      & \bestoverall{Graphs ICL (1 ex / 100 cls)} & \bestoverall{\best{30.25}} & \bestoverall{\best{26.93}} & \bestoverall{\best{11.25}} & \bestoverall{\best{40.56}} & \bestoverall{\best{40.40}} & \bestoverall{\second{20.24}} \\
      & Graphs ICL (2 ex / 100 cls)  & 29.90 & \second{26.90} & \second{11.15} & 38.79 & 39.87 & 19.96 \\
    
    \bottomrule
  \end{tabular}}
\end{table*}

\section{Implementation details}

All experiments use vLLM~\cite{vllm} for inference on up to four
NVIDIA L40S GPUs with greedy decoding (temperature $T{=}0$) and reasoning
modes disabled.  
Each clip is represented by $M{=}16$ uniformly sampled frames. We discard
the first and last frame, and build temporal windows over the remaining $14$ frames as follows. Windows begin at $f_2$, each anchored at its terminal frame, and grow incrementally with sizes $2, 3, 4$ until reaching the maximum size
$n{=}4$; thereafter, windows of size $n{=}4$ slide forward with stride
$s{=}1$. This produces $13$ overlapping windows in total ($2$ ramp-up windows and $11$ fixed-size windows). 
Graph construction therefore requires $2 \times 13 = 26$
VLM calls per clip (one Stage\,I and one Stage\,II per window), plus one
call for action prediction, totaling $27$ forward passes. Models that
fit within the four-GPU budget are run at native full precision; models that exceed it are run from the highest-precision publicly released
quantized checkpoint compatible with the budget (\cref{tab:vlms}).
Output responses that violate the JSON schema are re-prompted with the correction template for up to two attempts.
Samples that remain invalid after this loop are not discarded;
instead, we apply lenient post-processing in which invalid responses are removed and the remaining predictions are retained as a (possibly truncated) ranked list. All metrics are computed on the cleaned predictions. Samples for which cleaning leaves no valid prediction are counted as a miss at every rank.

\end{document}